\documentclass[11pt]{article}
\usepackage[letterpaper,margin=1in]{geometry}
\usepackage[T1]{fontenc}
\usepackage{lmodern}
\usepackage{graphicx}
\usepackage{booktabs}
\usepackage{amsmath,amssymb}
\usepackage{amsthm}
\usepackage{multirow}
\usepackage{xcolor}
\usepackage{xspace}
\usepackage{caption}

\usepackage[normalem]{ulem}
\usepackage{eccvabbrv}          
\usepackage[colorlinks=true,linkcolor=blue,citecolor=blue,urlcolor=blue]{hyperref}
\usepackage{cleveref}

\newtheorem{proposition}{Proposition}
\crefname{proposition}{Proposition}{Propositions}
\Crefname{proposition}{Proposition}{Propositions}
\AtBeginDocument{\def\doi#1{\url{https://doi.org/#1}}}
\providecommand{\keywords}[1]{\par\medskip\noindent
  {\def\and{\,\textperiodcentered\,}\textbf{Keywords:}~#1}}

\title{Spectral Gradient Orthogonalization Improves\\
       Differentially Private Training at Scale}
\author{Sabari Shanmugam \and Nick Barnes \and Kerry Taylor}
\date{\normalsize
  School of Computing, Australian National University, Canberra, Australia\\
  \texttt{\{sabari.shanmugam, nick.barnes, kerry.taylor\}@anu.edu.au}}

\begin{document}
\maketitle
\begingroup
\renewcommand\thefootnote{}%
\footnotetext{Accepted at the European Conference on Computer Vision (ECCV) 2026; to appear in Springer \emph{Lecture Notes in Computer Science}.}%
\addtocounter{footnote}{-1}\endgroup
\begin{abstract}
Differentially private training adds isotropic Gaussian noise to clipped gradients, corrupting every singular direction equally.
In vision models, where spatial correlation concentrates gradient energy into a low-rank subspace, most of this noise falls in directions that carry little signal.
Spectral gradient orthogonalization via polar decomposition is introduced as a post-processing step that recovers directional signal from the noisy gradient's low-rank structure at zero additional privacy cost.
A phase transition governs the utility of this approach: orthogonalization improves accuracy only when the per-direction spectral signal-to-noise ratio (SNR) suffices for singular vector recovery; in low-SNR regimes, the directional bias of the gradient is replaced by a nearly random orthogonal update, and the transformation is harmful.
The recovery threshold is determined by the spectral gap of the gradient and is surpassed at large batch sizes.
Empirically, the benefit scales with model capacity: spectral orthogonalization achieves a $+20.9\%$ improvement over DP-SGD on WRN-28-10 ($B = 4096$) and $+14.9\%$ on ResNet-18, while reducing inter-run variance by a factor of two to three.
In the fine-tuning regime, spectral orthogonalization matches the stability of DP-Adam while maintaining a first-order memory footprint.
Combining spectral with temporal denoising yields $50.3\%$ on CIFAR-10 ($\varepsilon = 4$), the highest accuracy in any tested configuration.
These gains are specific to moderate-to-high-SNR regimes such as large-batch training of higher-capacity models.
Small-batch or low-SNR settings are better served by DP-SGD or temporal denoising.

\keywords{Differential privacy \and Spectral methods \and Gradient orthogonalization \and Private training \and Polar decomposition}
\end{abstract}

\begin{figure}[t]
\centering
\includegraphics[width=\linewidth]{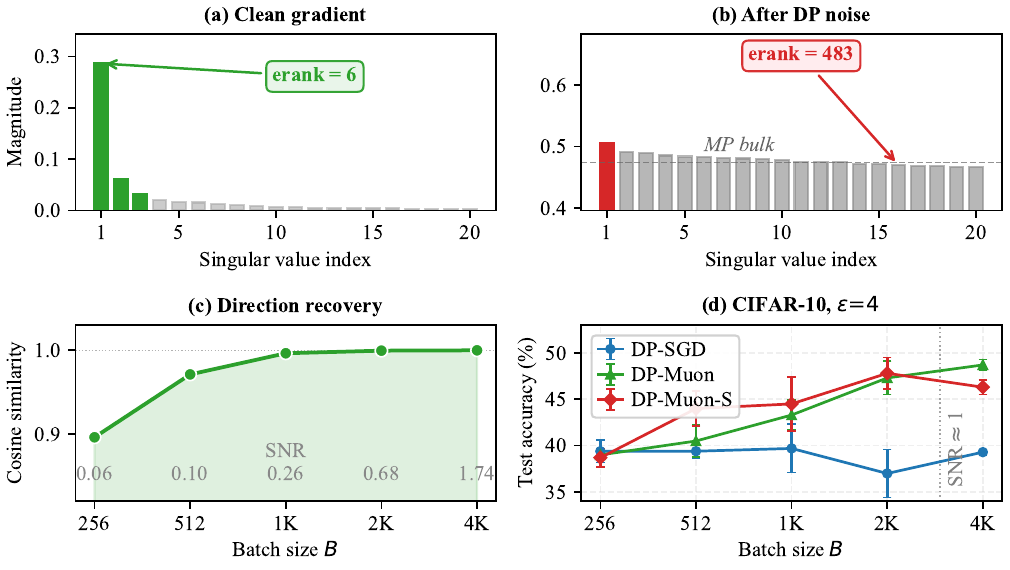}
\caption{Spectral gradient orthogonalization overview.
\emph{(a)}~Gradient singular values from a representative convolutional layer (epoch~1, CIFAR-10): most energy concentrates in a small number of directions (effective rank~$6$).
\emph{(b)}~After DP noise addition ($\varepsilon = 4$), isotropic perturbation raises the effective rank to $483$: the signal becomes indistinguishable from the noise-dominated singular values (Marchenko--Pastur bulk), making the noisy gradient nearly directionless.
\emph{(c)}~Direction recovery: cosine similarity between the leading singular vector of the noisy and clean gradients, measured via Monte Carlo simulation (100 trials per batch size) using the measured WRN-16-4 spectrum with noise calibrated to \cref{tab:snr}. High similarity indicates that the gradient's directional signal persists through the noise, providing the geometric condition for orthogonalization to succeed. Recovery transitions sharply from $0.90$ at $B\!=\!256$ to ${\geq}0.99$ at $B\!\geq\!1024$.
\emph{(d)}~Test accuracy on CIFAR-10 ($\varepsilon = 4$) follows the same pattern: DP-SGD stagnates while spectral methods scale with batch size past the recovery threshold.}
\label{fig:overview}
\end{figure}

\section{Introduction}
\label{sec:intro}
Computer vision models trained on sensitive data, such as medical images or visual biometrics (facial recognition, iris scans), must protect individual privacy while maintaining utility~\cite{kaissis2020secure}.
Differential privacy (DP)~\cite{dwork2006calibrating} provides a formal guarantee: the model's behavior changes negligibly whether or not any single example is included in the training set, preventing the memorization or recall of individual training samples.
The standard mechanism, differentially private stochastic gradient descent (DP-SGD)~\cite{abadi2016deep}, enforces this by clipping each per-sample gradient to a fixed norm and adding calibrated Gaussian noise before aggregation.
The noise ensures that the contribution of any one example is masked, but it degrades the optimization signal, reducing model accuracy.
This utility cost is especially severe when training from scratch, where every parameter must be learned under noise.
For vision architectures with large convolutional and attention layers, the gradient of each weight matrix is independently perturbed by isotropic noise, regardless of its structure.

Neural network training proceeds by computing gradients of the loss with respect to each weight matrix.
Recent work in optimization has shown that these gradients are approximately low-rank~\cite{li2018measuring,aghajanyan2021intrinsic}, a property exploited by optimizers such as Shampoo~\cite{gupta2018shampoo} and Muon~\cite{jordan2024muon} for faster convergence.
In vision models this structure is especially pronounced: the effective rank of a convolutional gradient is typically an order of magnitude smaller than the matrix dimension, a property we characterize empirically (\cref{sec:analysis}).
The noise added by DP-SGD, however, is \emph{spectrally indiscriminate}: isotropic Gaussian perturbation corrupts every singular direction equally, regardless of how much signal that direction carries.
The result is a spectral mismatch (\cref{fig:overview}b): the few directions that carry signal receive the prescribed noise, while the many directions that carry little signal become pure noise, yet contribute equally to the aggregate update.

While spectral orthogonalization is established in non-private optimization~\cite{jordan2024muon}, its utility under differential privacy is fundamentally regime-dependent.
Replacing the noisy gradient with its nearest orthogonal matrix $UV^\top$ via polar decomposition (computed efficiently through Newton-Schulz iteration) is shown to transition from harmful to substantially beneficial at a predictable SNR threshold, providing the first principled criterion for when spectral structure can be recovered from the noise bulk in private training.
By the post-processing theorem~\cite{dwork2014algorithmic}, this operation incurs zero additional privacy cost.
Orthogonalization preserves the gradient's singular vectors but sets all singular values to one, suppressing the inflated magnitudes of noise-dominated directions.
This is beneficial when the true signal directions are recoverable from the noisy gradient, but harmful otherwise: at low SNR, orthogonalization normalizes the noise bulk to unit norm, producing a nearly random update that is strictly worse than the noisy-but-directionally-biased gradient used by DP-SGD.

This leads to a theoretical prediction: orthogonalization should improve utility when the gradient signal-to-noise ratio (SNR) is high enough for the true singular vectors to be recoverable from the noisy gradient, and be neutral or harmful otherwise.
This prediction is verified empirically, revealing a \textbf{phase transition} governed by the SNR.
At large batch sizes, the noise per element shrinks inversely with $B$, pushing the SNR past the recovery threshold.
The crossover batch size depends on task complexity: it occurs at $B \approx 1024$ on CIFAR-10, shifts to $B \approx 4096$ to $8192$ on CIFAR-100 where the gradient signal is ${\sim}9\times$ weaker, and is verified across three datasets in \cref{sec:batch_sweep}.

This phenomenon is characterized through a systematic empirical study spanning batch sizes $B \in \{256, 512, \ldots, 8192\}$ and privacy budgets $\varepsilon \in \{2, 3, 4, 6, 8\}$.
The contributions of this work are three-fold.
First, a recovery condition (\cref{prop:recovery}) linking batch size, noise multiplier, and spectral gap is derived and verified across CIFAR-10, CIFAR-100, and SVHN, revealing that spectral orthogonalization transitions from harmful to substantially beneficial as the gradient SNR crosses a predictable threshold.
Second, spectral gradient orthogonalization via approximate polar decomposition (Newton-Schulz iteration) is introduced as a deterministic post-processing step that preserves the $(\varepsilon, \delta)$-DP guarantee at zero additional privacy cost, achieving $+20.9\%$ over DP-SGD on WRN-28-10 ($B = 4096$, $\varepsilon = 4$) and reducing inter-run variance by a factor of two to three.
Third, the complementarity of spectral and temporal denoising is demonstrated: combining orthogonalization with low-pass filtering (DOPPLER) yields $50.3\%$ on CIFAR-10 ($\varepsilon = 4$), the highest accuracy in any tested configuration.

\section{Related Work}
\label{sec:related}
\paragraph{Differentially private optimization.}
DP-SGD~\cite{abadi2016deep} provides $(\varepsilon, \delta)$-DP guarantees via per-sample gradient clipping and Gaussian noise addition.
Subsequent work has improved the utility-privacy trade-off through better accounting~\cite{mironov2017renyi,balle2020hypothesis}, larger batch training~\cite{de2022unlocking}, and architectural choices~\cite{ponomareva2023dpfy}.
DP-Adam~\cite{kingma2015adam} applies adaptive learning rates to privatized gradients but has shown limited benefit at moderate privacy budgets in our experiments, consistent with findings that adaptive methods require careful tuning under DP~\cite{bu2023dpoptimization}.

\paragraph{Noise reduction via spectral and temporal filtering.}
A growing line of work reduces the impact of DP noise through spectral or temporal post-processing.
Spectral-DP~\cite{feng2023spectral} adds calibrated noise in the frequency domain rather than the spatial domain and applies spectral filtering, achieving the same $(\varepsilon,\delta)$-DP guarantee with lower effective perturbation.
DOPPLER~\cite{zhang2024doppler} treats the privatized gradient sequence as a noisy signal and applies a low-pass filter to suppress high-frequency noise components while preserving low-frequency gradient information, yielding three to ten percent accuracy improvements across datasets and models.
DiSK~\cite{zhang2025disk} extends this direction with a simplified Kalman filter that adapts its gain over training, achieving best-reported from-scratch results on CIFAR-100 and ImageNet-1k using augmentation multiplicity and architecture-specific tuning.
Other approaches include frequency-domain gradient computation (GReDP~\cite{wang2024gredp}) and joint spectral-temporal filtering (FFTKF~\cite{yun2025fftkf}).

These methods modify the privacy mechanism or its inputs.
Spectral-DP and GReDP shape where noise enters by operating in the frequency domain, while DOPPLER and DiSK filter the privatized gradient across the training trajectory.
Anisotropic Gaussian mechanisms~\cite{ji2024lessismore} take a related route by reshaping the noise covariance at injection, which requires estimating that covariance from public data or a separate privacy budget and modifying the accounting.
Spectral orthogonalization differs on the axis that is central to this work.
It leaves the standard isotropic mechanism untouched and applies only deterministic post-processing to the already-privatized gradient, recovering directional signal from its \emph{low-rank matrix structure} at zero additional privacy cost, with no covariance estimation and no change to the accounting.
Because it targets matrix structure rather than component-wise magnitudes, it is complementary to the frequency-domain and temporal methods above (\cref{sec:temporal_comparison}).

\paragraph{Parameter-efficient private fine-tuning.}
Parameter-efficient methods 
(e.g., DP-BiTFiT~\cite{bu2022dpbitfit}, low-rank reparametrization~\cite{yu2021differentially}, and LoRA~\cite{hu2022lora,li2021large}) reduce trainable parameters to lower per-parameter noise.
In this regime, all optimizers achieve similar accuracy, consistent with the finding that low-rank constraints remove the spectral mismatch that orthogonalization addresses in full-parameter training.

\paragraph{Gradient structure and low-rank methods.}
Neural network gradients exhibit low-rank structure~\cite{li2018measuring,aghajanyan2021intrinsic}: most gradient energy concentrates in a small subspace whose dimension correlates with task complexity.
GaLore~\cite{zhao2024galore} projects gradients into a low-rank subspace for memory efficiency.
Our work differs in applying spectral processing as a \emph{post-processing} step on the already-privatized gradient, preserving the DP guarantee by the post-processing theorem~\cite{dwork2014algorithmic}.

\paragraph{Spectral methods in optimization.}
Shampoo~\cite{gupta2018shampoo}, SOAP~\cite{vyas2024soap}, and Muon~\cite{jordan2024muon} exploit spectral structure for preconditioning or orthogonalization in non-private training.
Their interaction with DP noise injection has not been studied; spectral gradient orthogonalization is investigated here in the DP setting, identifying the batch-size regime in which it provides a benefit.

\section{Method}
\label{sec:method}

\subsection{Preliminaries}

\paragraph{Differential Privacy.}
A randomized mechanism $\mathcal{M}$ satisfies $(\varepsilon, \delta)$-differential privacy~\cite{dwork2006calibrating} if for all adjacent datasets $D, D'$ differing in a single record and all measurable sets $S$:
$\Pr[\mathcal{M}(D) \in S] \leq e^{\varepsilon} \Pr[\mathcal{M}(D') \in S] + \delta$.

\paragraph{DP-SGD.}
Each per-sample gradient $g_i$ is clipped to bound its Frobenius norm: $\bar{g}_i = g_i \cdot \min(1, C / \|g_i\|_F)$, where $C$ is the clipping threshold.
The clipped gradients are averaged and perturbed:
\begin{equation}
    \tilde{G} = \frac{1}{B}\left(\sum_{i=1}^{B} \bar{g}_i + \mathcal{N}(0, \sigma^2 C^2 \mathbf{I})\right),
    \label{eq:dpsgd}
\end{equation}
where $\sigma$ is the noise multiplier calibrated via R\'enyi DP accounting~\cite{mironov2017renyi}.

\paragraph{Post-Processing Theorem.}
Any function of the output of a differentially private mechanism preserves the same $(\varepsilon, \delta)$ guarantee~\cite{dwork2014algorithmic}.
This is the key property enabling our approach: all spectral processing is applied \emph{after} 
noise addition. 

\subsection{Spectral Gradient Orthogonalization}
\label{sec:spectral}

For a 2D parameter tensor, the method proceeds as:

\paragraph{Step 0: DP-SGD Gradient.}
Per-sample gradients are clipped and noised exactly as in standard DP-SGD (\cref{eq:dpsgd}), producing the privatized gradient $\tilde{G} \in \mathbb{R}^{m \times n}$.
All subsequent steps operate on $\tilde{G}$ and are therefore post-processing.

\paragraph{Step 1: Momentum Accumulation.}
Following the Muon formulation~\cite{jordan2024muon}, EMA-style Nesterov momentum is applied to the privatized gradient $\tilde{G}$:
$M_t = \beta M_{t-1} + (1 - \beta) \tilde{G}_t$, $\hat{M}_t = (1-\beta) \tilde{G}_t + \beta M_t$,
where $\beta$ is the momentum coefficient and $\hat{M}_t$ is the Nesterov lookahead.
The $(1 - \beta)$ scaling produces a weighted average rather than an accumulation, following the original Muon implementation.

\paragraph{Step 2: Orthogonalization via Newton-Schulz.}
The momentum matrix is orthogonalized using five iterations of a quintic Newton-Schulz scheme~\cite{jordan2024muon}:
\begin{equation}
    X_0 = \hat{M}_t / \|\hat{M}_t\|_F, \quad
    X_{k+1} = aX_k + bX_kX_k^\top X_k + cX_k(X_kX_k^\top)^2,
\end{equation}
where $(a, b, c) = (3.4445, -4.7750, 2.0315)$ are coefficients optimized for convergence speed.
Frobenius-norm initialization ensures all singular values of $X_0$ are at most one (since $\|X_0\|_2 \leq \|X_0\|_F = 1$), which is sufficient for Newton-Schulz convergence.
The output $X_K \approx UV^\top$, the closest orthogonal matrix to $\hat{M}_t$.
The iterations require only matrix multiplications.
The resulting wall-clock and memory overhead is modest and is measured per configuration in 
supp material.

\paragraph{Step 3: Dimension Scaling. }A correction factor $\hat{G} = \sqrt{\max(1, m/n)} \cdot X_K$ accounts for rectangular gradients.\footnote{
We cap
scaling factor 
to $4.0$ to prevent instability for highly rectangular matrices.}
The parameter update is $W_{t+1} = W_t - \eta \hat{G}$.
For 1D parameters (biases, layer norms), standard DP-SGD with momentum is used.

\subsection{Magnitude-Preserving Variant and Privacy}
\label{sec:scaled}

Pure orthogonalization discards all singular value information.
At higher SNR the leading singular value carries a meaningful magnitude signal, so a variant rescales the orthogonal update:
\begin{equation}
    \hat{G}_{\text{scaled}} = \sigma_1(M_t) \cdot \sqrt{\max\!\left(1, \frac{m}{n}\right)} \cdot X_K,
    \label{eq:scaled}
\end{equation}
where $M_t = \beta M_{t-1} + \tilde{G}_t$ is a heavy-ball momentum buffer (without Nesterov lookahead), $X_K$ is the Newton-Schulz output applied to $M_t$, and $\sigma_1(M_t)$ is its spectral norm.
Since $\sigma_1(M_t)$ is computed from the noisy buffer, it overestimates the clean magnitude by up to $\|N\|_2$, a bias that diminishes as the SNR increases.
Both variants apply only deterministic post-processing to the Gaussian mechanism output (\cref{eq:dpsgd}); by the post-processing theorem, the privacy guarantee is identical to DP-SGD with no additional cost.

In practice, the magnitude-preserving variant is recommended only at moderate-to-high SNR, where the spectral-norm estimate $\sigma_1(M_t)$ is reliable.
At low SNR 
magnitude bias dominates, and pure orthogonalization is preferable.
It is also not recommended on SVHN where it shows unstable bimodal convergence.

\section{Why Orthogonalization Helps at Large Batch Sizes}
\label{sec:analysis}

\subsection{Signal-to-Noise Ratio in DP Gradients}

The noisy gradient~\eqref{eq:dpsgd} decomposes as $\tilde{G} = G + N$, where $G = \frac{1}{B}\sum_i \bar{g}_i$ is the clipped average and $N \sim \mathcal{N}(0, \frac{\sigma^2 C^2}{B^2} \mathbf{I})$ is the privacy noise.
The noise $N$ is i.i.d.\ Gaussian by construction of the Gaussian mechanism~\cite{abadi2016deep,dwork2006calibrating} and is independent of the data; clipping affects the signal $G$ but not the noise distribution.
The per-element noise standard deviation scales as $\sigma C / B$: \emph{larger batches reduce the noise magnitude relative to the signal}.

The signal $G$ is approximately low-rank, with effective rank~\cite{roy2007effective} $\mathrm{erank}(G) = (\sum_i \sigma_i(G))^2 / \sum_i \sigma_i(G)^2 \ll \min(m, n)$.
After adding DP noise, the effective rank increases dramatically (\eg, from $\sim\!11$ to $\sim\!197$ for WRN-16-4 at $\varepsilon = 4$), as noise fills all singular directions.

\begin{figure}[t]
\centering
\includegraphics[width=\linewidth]{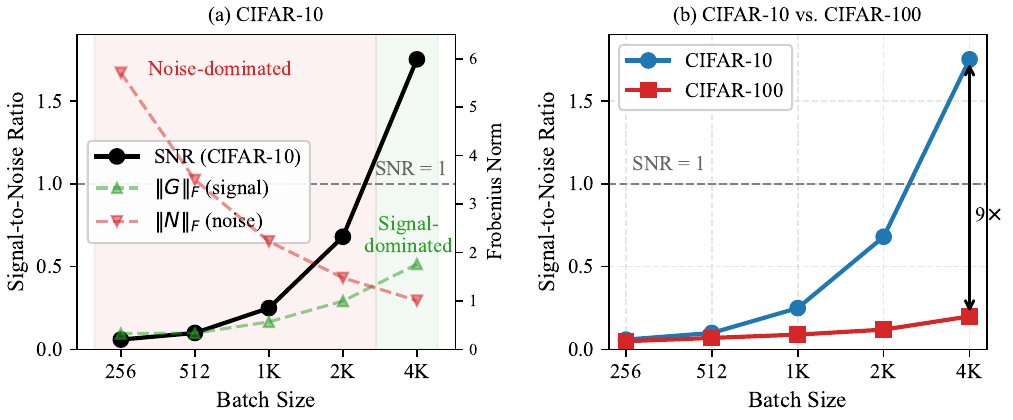}
\caption{Gradient SNR vs.\ batch size at $\varepsilon = 4$. \emph{(a)}~CIFAR-10 SNR with signal and noise Frobenius norms; the horizontal dashed line marks $\text{SNR} = 1$, above which orthogonalization recovers useful gradient structure. \emph{(b)}~CIFAR-10 vs.\ CIFAR-100: the SNR gap widens with batch size, reaching ${\sim}9\times$ at $B = 4096$.}
\label{fig:snr}
\end{figure}

\cref{tab:snr,fig:snr} report measured SNR values across batch sizes for WRN-16-4 at $\varepsilon = 4$.
On CIFAR-10, the SNR increases from $0.06$ at $B = 256$ to $1.75$ at $B = 4096$, crossing $1.0$ between $B = 2048$ and $B = 4096$.
On CIFAR-100, the SNR is substantially lower at every batch size ($0.20$ vs $1.75$ at $B = 4096$; $0.12$ vs $0.68$ at $B = 2048$), because the gradient signal norm \emph{decreases} with batch size on this harder task rather than growing as on CIFAR-10.
The Frobenius SNR is a conservative proxy for the spectral recovery condition: since $\|N\|_F / \|N\|_2 \approx \sqrt{\min(m,n)}$, a Frobenius SNR of $1$ implies $\sigma_1(G) > \|N\|_2$ by a factor of $\sqrt{\min(m,n)}$, so the actual recovery threshold is reached at smaller batch sizes than the $\text{SNR} = 1$ line suggests in \cref{fig:snr}.
In particular, orthogonalization can recover the leading singular direction even when the Frobenius SNR is below $1$, provided the gradient signal is concentrated in a spiked leading direction whose magnitude exceeds the noise operator norm $\|N\|_2$.

\begin{table}[t]
\centering
\caption{Measured gradient SNR ($\|G\|_F / \|N\|_F$) for WRN-16-4 at $\varepsilon = 4$, averaged over the first epoch. $\sigma$ is the noise multiplier (identical for both datasets at $N = 50\text{K}$). Signal and noise norms are shown for CIFAR-10. As training progresses, gradient structure typically becomes more low-rank, which lowers the recovery threshold.}\label{tab:snr}
\begin{tabular}{c c c c c c}
\toprule
$B$ &~~~$\sigma$~~~&~~~$\|G\|_F$ (signal)~~~&~~~$\|N\|_F$ (noise)~~~&~~~SNR (C-10)~~~& SNR (C-100) \\
\midrule
256  & 0.88 & 0.33 & 5.71 & 0.06 & 0.05 \\
512  & 1.08 & 0.34 & 3.50 & 0.10 & 0.07 \\
1024 & 1.38 & 0.57 & 2.23 & 0.25 & 0.09 \\
2048 & 1.83 & 1.00 & 1.48 & 0.68 & 0.12 \\
4096 & 2.49 & 1.76 & 1.01 & $\mathbf{1.75}$ & 0.20 \\
\bottomrule
\end{tabular}
\end{table}

\subsection{Spectral Recovery Threshold}

The spectral-gap and SNR measurements used in this section (\cref{tab:snr}) are obtained from a separate post-hoc analysis run.
They are not released and are not used during training, hyperparameter selection, or model release.
Deployed models are accounted for under standard DP-SGD, with no contribution from these statistics.
The recovery threshold below is therefore an explanatory framework, not a private estimator for selecting batch size or enabling orthogonalization.

Orthogonalization replaces $\tilde{G}$ with $UV^\top$, projecting onto the Stiefel manifold~\cite{edelman1998geometry}.
This transformation preserves directions but normalizes magnitudes.
The effect depends on the SNR:

\textbf{Low SNR (small batch, small $\varepsilon$):} The noise dominates. $UV^\top$ reflects the singular structure of the noise rather than the signal. Orthogonalization amplifies noisy directions to unit norm, providing no benefit.

\textbf{High SNR (large batch, large $\varepsilon$):} The signal's top singular directions persist through the noise.
By the Wedin $sin\ \theta$ theorem~\cite{wedin1972perturbation}, the angle $\theta$ between the clean and noisy leading singular vectors satisfies $\sin\theta \leq \|N\|_2 / (\sigma_1(G) - \sigma_2(G))$.
When the spectral gap exceeds the noise magnitude, orthogonalization approximately recovers the signal directions.

The observed threshold behavior is conceptually analogous to the phase transitions studied in spiked matrix models~\cite{baik2005phase}, where signal recovery occurs only when the dominant singular value exceeds the noise operator norm.
The Wedin $\sin\theta$ bound provides a rigorous recovery condition without requiring the distributional assumptions of the spiked model.

\begin{proposition}[Recovery condition]
\label{prop:recovery}
Let $\tilde{G} = G + N \in \mathbb{R}^{m \times n}$ with $m \leq n$, where $G$ has singular values $\sigma_1(G) \geq \sigma_2(G) \geq 0$ and $N$ has i.i.d.\ entries with variance $\sigma^2 C^2 / B^2$.
By the Wedin $\sin\theta$ bound~\cite{wedin1972perturbation}, the angle $\theta$ between the leading left singular vectors of $\tilde{G}$ and $G$ satisfies:
\begin{equation}
    \sin\theta \leq \frac{\|N\|_2}{\sigma_1(G) - \sigma_2(G)}.
    \label{eq:wedin}
\end{equation}
Since $\|N\|_2 \leq \sigma C (\sqrt{m} + \sqrt{n}) / B$ with high probability~\cite{vershynin2012introduction}, orthogonalization recovers the signal subspace when:
\begin{equation}
    B > \frac{\sigma C \left(\sqrt{m} + \sqrt{n}\right)}{\sigma_1(G) - \sigma_2(G)}.
    \label{eq:batch_threshold}
\end{equation}
\end{proposition}

\noindent The recovery condition is derived by applying the Wedin $\sin\theta$ bound~\cite{wedin1972perturbation,vu2011singular} to the scaling laws of the Gaussian mechanism; a step-by-step derivation linking each bound to its source theorem is provided in the supplementary material, Section~1.
The spectral gap $\sigma_1(G) - \sigma_2(G)$ is measured at epoch~1, the hardest regime for recovery since gradient structure becomes more low-rank as training progresses~\cite{li2018measuring}; the threshold is therefore conservative.
\Cref{eq:batch_threshold} yields a minimum batch size that depends on the noise multiplier $\sigma$, the clipping threshold $C$, the layer dimensions $m, n$, and the spectral gap of the clean gradient.
Datasets with weaker gradient signal (\ie smaller spectral gap) require proportionally larger batches, consistent with the ${\sim}9\times$ SNR gap between CIFAR-10 and CIFAR-100 (\cref{tab:snr}) and the corresponding shift in the crossover batch size observed in \cref{sec:batch_sweep}.
\Cref{prop:recovery} provides a necessary condition for leading singular vector recovery; full subspace recovery requires analogous gaps at each rank, which is verified empirically via the effective rank measurements in \cref{tab:snr}.

\paragraph{Numerical validation.}
Plugging measured spectral gaps from ResNet-18~\cite{he2016resnet} (epoch~1, CIFAR-10) into \cref{eq:batch_threshold} with the noise multipliers from \cref{tab:snr}, most layers satisfy the recovery condition already at $B = 256$, and all layers do by $B = 512$ (per-layer thresholds in \cref{tab:bstar}).
The network-wide bottlenecks are the deepest convolutions ($512 \times 4608$, spectral gap $\Delta\sigma \approx 0.27$, $\sqrt{m} + \sqrt{n} \approx 90$) and layer3.1.conv1 ($256 \times 2304$, $\Delta\sigma \approx 0.14$), which transition between $B = 256$ and $B = 512$.
This aligns with the empirical crossover for ResNet-18 in \cref{fig:results_grid}(b), where DP-Muon already leads DP-SGD by $+5.8\%$ at $B = 256$ (when most layers have recovered) and the gap widens steadily with batch size.
For the wider WRN-16-4, increased layer dimensionality raises $\sqrt{m} + \sqrt{n}$ and pushes the predicted $B^*$ higher, consistent with the observed crossover at $B \approx 2048$.

This aligns with the observed phase transition (\cref{tab:snr}): at $B \leq 512$ ($\text{SNR} < 0.1$) orthogonalization amplifies noise; at $B = 4096$ ($\text{SNR} = 1.75$) the signal dominates and improvements are consistent.
DP-SGD stagnates because it uses the noisy gradient as-is, gaining little from improved directional structure; orthogonalization converts that structure into a cleaner update, explaining why spectral methods scale faster with both $B$ and $\varepsilon$ (\cref{fig:eps_sweep}).

\subsection{Comparison with Temporal Filtering}
\label{sec:temporal_comparison}

Recent methods such as DOPPLER~\cite{zhang2024doppler} and DiSK~\cite{zhang2025disk} reduce DP noise by applying temporal filters to the sequence of privatized gradients, also as post-processing steps incurring no additional privacy cost.
The two classes of methods scale differently with batch size: temporal filtering provides roughly constant noise reduction determined by the filter coefficient $\beta$, largely independent of batch size.
Spectral orthogonalization, by contrast, depends on the per-step SNR and grows more effective with batch size as the spectral gap exceeds the noise operator norm (\cref{sec:analysis}).
At small batch sizes temporal filtering dominates; at large batch sizes spectral orthogonalization catches up and eventually dominates.
Since the two target different noise structures (temporal averaging vs.\ spatial structure), they are complementary and can be combined (\cref{sec:temporal}).

\section{Experiments}
\label{sec:experiments}

\subsection{Experimental Setup}
\label{sec:setup}

\paragraph{Architecture and Datasets.}
Wide ResNet (WRN-16-4)~\cite{zagoruyko2016wrn} is used for all from-scratch experiments.
Three datasets are evaluated: CIFAR-10 and CIFAR-100~\cite{krizhevsky2009cifar} (50K training samples each) and SVHN~\cite{netzer2011svhn} (73K training samples).
Per-sample Frobenius norm clipping with $C = 1.0$ and $\delta = 10^{-5}$ is used throughout.
The noise multiplier $\sigma$ is calibrated via R\'enyi DP accounting~\cite{mironov2017renyi,yousefpour2021opacus} to achieve the target $\varepsilon$.
Training epochs are held fixed across batch sizes; larger batches therefore take fewer gradient steps per epoch.

\paragraph{Methods.}
Eight methods are compared, all sharing the same per-sample clipping and noise addition:
(i)~\textbf{DP-SGD} (momentum $\beta\!=\!0.9$, LR$\,{=}\,0.3$);
(ii)~\textbf{DP-Adam}~\cite{kingma2015adam} ($\beta_1\!=\!0.9$, $\beta_2\!=\!0.999$, LR$\,{=}\,0.001$);
(iii)~\textbf{DP-Muon} ($UV^\top$ orthogonalization, Nesterov $\beta\!=\!0.95$, LR$\,{=}\,0.02$);
(iv)~\textbf{DP-Muon-S} ($\sigma_1 \cdot UV^\top$, \cref{eq:scaled}, heavy-ball $\beta\!=\!0.9$, LR$\,{=}\,0.3$);
(v)~\textbf{DiSK-SGD}~\cite{zhang2025disk} (Kalman filter, $\kappa\!=\!1.5$, on DP-SGD);
(vi)~\textbf{DiSK-Muon} (Kalman filter on DP-Muon);
(vii)~\textbf{DOPPLER-SGD}~\cite{zhang2024doppler} (EMA low-pass on DP-SGD);
(viii)~\textbf{DOPPLER-Muon} (EMA low-pass on DP-Muon).
All methods use cosine learning rate decay with linear warmup.
Results report best test accuracy (\%) across epochs, averaged over three seeds (42, 43, 44) with standard deviation.
DP-Muon uses $\beta = 0.95$ following the original Muon implementation~\cite{jordan2024muon}; the orthogonalization step, not the momentum coefficient, drives the gain, as DP-Muon-S uses identical $\beta = 0.9$ heavy-ball momentum to DP-SGD yet achieves comparable improvements.

\subsection{Batch Size Sweep}
\label{sec:batch_sweep}

\begin{figure}[t]
\centering
\includegraphics[width=0.95\linewidth]{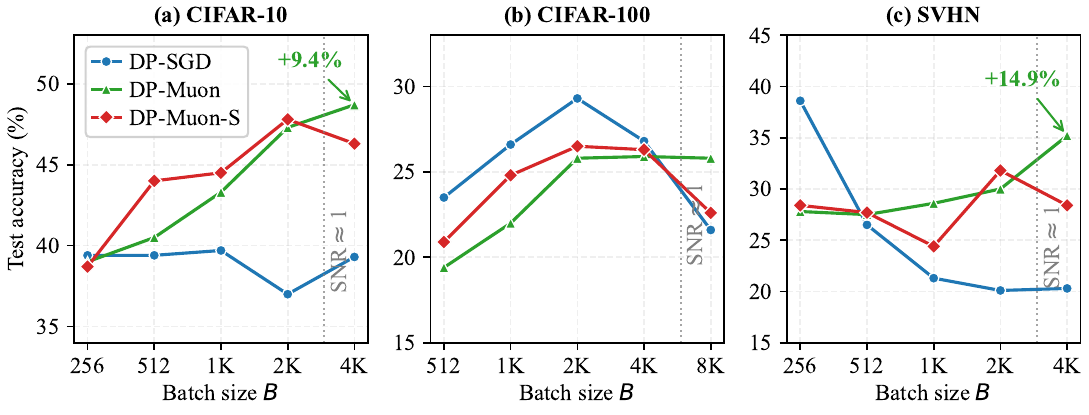}
\caption{Test accuracy vs.\ batch size at $\varepsilon = 4$ across three datasets.
\emph{(a)}~CIFAR-10: DP-SGD is flat while spectral methods scale upward ($+9.4\%$ at $B = 4096$).
\emph{(b)}~CIFAR-100: the crossover shifts to larger batch sizes, consistent with the lower SNR.
\emph{(c)}~SVHN: DP-SGD degrades with batch size while DP-Muon improves ($+14.9\%$ at $B = 4096$).
Vertical dotted lines mark the approximate $\text{SNR} \approx 1$ threshold. Mean over three seeds.}
\label{fig:batch_sweep}
\end{figure}

Batch size is varied at a fixed privacy budget of $\varepsilon = 4$.
\cref{tab:cifar10_batch,tab:cifar100_batch} report results on CIFAR-10/100;
\cref{fig:batch_sweep} visualizes the trends across all three datasets.

\begin{table}[t]
\centering
\caption{Test accuracy (\%) on CIFAR-10 at $\varepsilon = 4$. Mean $\pm$ std over three seeds. Best per column in \textbf{bold}.}\label{tab:cifar10_batch}
\begin{tabular}{l c c c c c}
\toprule
Method & $B\!=\!256$ & $B\!=\!512$ & $B\!=\!1024$ & $B\!=\!2048$ & $B\!=\!4096$ \\
\midrule
DP-SGD    & $39.4 \pm 1.2$           & $39.4 \pm 0.7$           & $39.7 \pm 2.6$           & $37.0 \pm 2.6$           & $39.3 \pm 0.3$ \\
DP-Adam   & $\mathbf{39.9 \pm 2.9}$  & $41.0 \pm 1.1$           & $41.8 \pm 1.8$           & $42.0 \pm 0.5$           & $38.8 \pm 1.7$ \\
DP-Muon   & $39.0 \pm 0.5$           & $40.5 \pm 1.7$           & $43.3 \pm 1.6$           & $47.3 \pm 1.8$           & $\mathbf{48.7 \pm 0.6}$ \\
DP-Muon-S & $38.7 \pm 1.0$           & $\mathbf{44.0 \pm 1.9}$  & $\mathbf{44.5 \pm 2.9}$  & $\mathbf{47.8 \pm 1.7}$  & $46.3 \pm 0.8$ \\
\bottomrule
\end{tabular}
\end{table}

\noindent\textbf{CIFAR-10:} At $B = 256$, all methods perform within $1.3\%$ of each other, consistent with the low-SNR regime ($\text{SNR} = 0.06$, \cref{tab:snr}).
DP-SGD remains flat across batch sizes ($37.0$ to $39.7\%$), while spectral methods scale upward: DP-Muon reaches $48.7\%$ at $B = 4096$ ($+9.4\%$).
DP-Muon-S peaks at $B = 2048$ ($47.8\%$) and declines at $B = 4096$, suggesting magnitude preservation is less beneficial at high SNR.
DP-Adam shows no batch-size scaling~\cite{bu2023dpoptimization}.

\begin{table}[t]
\centering
\caption{Test accuracy (\%) on CIFAR-100 at $\varepsilon = 4$ with tuned learning rates (DP-SGD $\eta\!=\!1.0$, DP-Adam $\eta\!=\!0.01$, DP-Muon $\eta\!=\!0.1$, DP-Muon-S $\eta\!=\!0.5$). Mean $\pm$ std over three seeds. Best per column in \textbf{bold}.}\label{tab:cifar100_batch}
\begin{tabular}{l c c c c c}
\toprule
Method & $B\!=\!512$ & $B\!=\!1024$ & $B\!=\!2048$ & $B\!=\!4096$ & $B\!=\!8192$ \\
\midrule
DP-SGD    & $23.5 \pm 1.0$           & $26.6 \pm 2.6$           & $\mathbf{29.3 \pm 1.2}$  & $\mathbf{26.8 \pm 0.6}$  & $21.6 \pm 0.9$ \\
DP-Adam   & $\mathbf{25.5 \pm 1.3}$  & $\mathbf{27.4 \pm 1.8}$  & $28.9 \pm 0.7$           & $26.0 \pm 0.7$           & $18.6 \pm 1.2$ \\
DP-Muon   & $19.4 \pm 1.7$           & $22.0 \pm 1.0$           & $25.8 \pm 0.7$           & $25.9 \pm 0.5$           & $\mathbf{25.8 \pm 0.6}$ \\
DP-Muon-S & $20.9 \pm 1.2$           & $24.8 \pm 0.3$           & $26.5 \pm 0.7$           & $26.3 \pm 0.9$           & $22.6 \pm 1.0$ \\
\bottomrule
\end{tabular}
\end{table}

\noindent\textbf{CIFAR-100:} The gradient SNR is ${\sim}9\times$ lower than on CIFAR-10 (\cref{tab:snr}), so \cref{prop:recovery} predicts a proportionally larger batch is needed.
At $B \leq 1024$, DP-Muon trails DP-SGD by $3$ to $5\%$: in this low-SNR regime, orthogonalization replaces a noisy but directionally biased gradient with a nearly random orthogonal update, which is strictly worse (as formalized by the recovery condition of \cref{prop:recovery}).
Direction recovery (\cref{fig:overview}c) exceeds $0.99$ only at $B = 4096$ (vs.\ $B = 1024$ on CIFAR-10), yet the accuracy crossover lags because orthogonalization normalizes noise-only directions (the Marchenko--Pastur bulk~\cite{marchenko1967distribution}) to unit norm alongside recovered signal directions.
With tuned learning rates, DP-SGD peaks at $B = 2048$ ($29.3\%$) and declines to $21.6\%$ at $B = 8192$, while DP-Muon increases steadily to $25.8\%$, overtaking by $+4.2\%$ at the largest batch size.

\noindent\textbf{SVHN:} DP-SGD accuracy decreases monotonically (\cref{fig:batch_sweep}c) from $38.6\%$ at $B=256$ to $20.3\%$ at $B=4096$.
DP-Muon increases by $14.9\%$ from $27.8\%$ to $35.2\%$
at the largest batch size.
We omit DP-Muon-S due to bimodal convergence. 

\noindent\textbf{Cross-Dataset Summary:} The crossover batch size shifts with SNR: $B \approx 1024$ on CIFAR-10, $B \approx 4096$ to $8192$ on CIFAR-100, and $B \approx 256$ on SVHN, consistent with \cref{prop:recovery}.
DiSK~\cite{zhang2025disk} reports $42.0\%$ on CIFAR-100 ($\varepsilon = 8$) using augmentation multiplicity and WRN-28-10; isolating the interaction with spectral processing is deferred to future work.

\subsection{Privacy Budget Sweep}
\label{sec:eps_sweep}

\begin{figure}[t]
\centering
\includegraphics[width=0.95\linewidth]{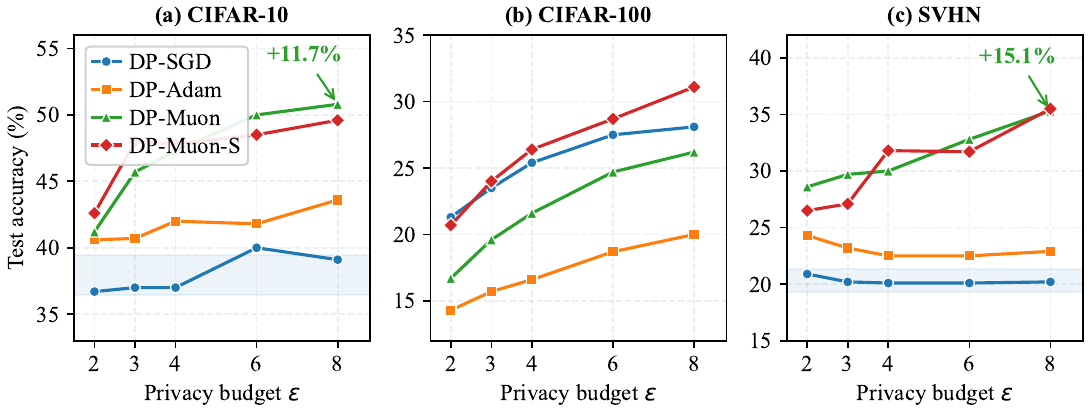}
\caption{Test accuracy vs.\ $\varepsilon$ at $B = 2048$ across three datasets. DP-SGD stagnates on CIFAR-10 and SVHN (shaded bands) while spectral methods scale with the privacy budget. DP-Muon-S is strongest on CIFAR-100 at all $\varepsilon$. Mean over three seeds.}
\label{fig:eps_sweep}
\end{figure}

\Cref{fig:eps_sweep} shows accuracy at $B = 2048$ across privacy budgets $\varepsilon \in \{2, 3, 4, 6, 8\}$.
DP-Muon accuracy scales steadily with the privacy budget (from $41.2\%$ at $\varepsilon=2$ to $50.8\%$ at $\varepsilon=8$ on CIFAR-10), while DP-SGD remains largely flat ($36.7$ to $40.0\%$).
On CIFAR-100, DP-Muon-S outperforms DP-SGD at $\varepsilon \geq 3$, with the advantage growing to $+3.0\%$ at $\varepsilon=8$.
On SVHN, DP-SGD is flat at ${\sim}20\%$ regardless of $\varepsilon$, while DP-Muon improves from $28.6\%$ to $35.3\%$.

\subsection{Comparison with Temporal Denoising}
\label{sec:temporal}

Recent temporal denoising methods, DiSK~\cite{zhang2025disk} (Kalman filtering) and DOPPLER~\cite{zhang2024doppler} (low-pass filtering), reduce DP noise by filtering the sequence of privatized gradients across training steps.
As discussed in \cref{sec:temporal_comparison}, temporal and spectral filtering target different noise reduction mechanisms and are in principle complementary.
\begin{figure}[t]
\centering
\includegraphics[width=0.68\linewidth]{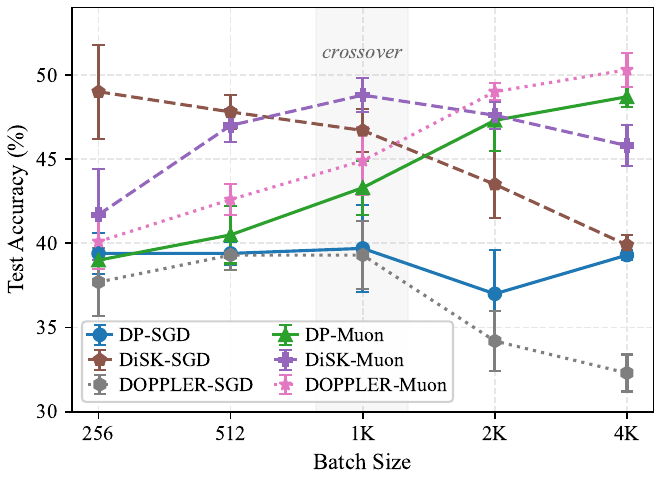}
\caption{Temporal vs.\ spectral denoising on CIFAR-10 at $\varepsilon = 4$. Temporal methods (DiSK-SGD, dashed brown) dominate at small batch sizes but decline as fewer gradient steps reduce the filter's averaging window. Spectral methods (DP-Muon, solid green) improve with batch size as per-step SNR increases. DOPPLER-Muon (dotted pink) combines both mechanisms and reaches the highest accuracy ($50.3\%$) at $B = 4096$. Error bars show $\pm$1 std over three seeds.}
\label{fig:temporal}
\end{figure}
\begin{table}[t]
\centering
\caption{Temporal vs.\ spectral denoising on CIFAR-10 ($\varepsilon = 4$). DiSK-SGD applies the Kalman filter~\cite{zhang2025disk} ($\kappa = 1.5$) to DP-SGD. DOPPLER-Muon combines low-pass filtering~\cite{zhang2024doppler} with spectral orthogonalization. Mean $\pm$ std over three seeds.}\label{tab:temporal}
{\small
\begin{tabular}{l l c c c c c}
\toprule
Denoising & Method & $B\!=\!256$ & $B\!=\!512$ & $B\!=\!1024$ & $B\!=\!2048$ & $B\!=\!4096$ \\
\midrule
None     & DP-SGD        & $39.4 \pm 1.2$ & $39.4 \pm 0.7$ & $39.7 \pm 2.6$ & $37.0 \pm 2.6$ & $39.3 \pm 0.3$ \\
Temporal & DiSK-SGD      & $\mathbf{49.0 \pm 2.8}$ & $\mathbf{47.8 \pm 1.0}$ & $46.7 \pm 1.3$ & $43.5 \pm 2.0$ & $39.9 \pm 0.6$ \\
Temporal & DOPPLER-SGD   & $37.7 \pm 2.0$ & $39.3 \pm 0.9$ & $39.3 \pm 2.0$ & $34.2 \pm 1.8$ & $32.3 \pm 1.1$ \\
\midrule
Spectral & DP-Muon       & $39.0 \pm 0.5$ & $40.5 \pm 1.7$ & $43.3 \pm 1.6$ & $47.3 \pm 1.8$ & $48.7 \pm 0.6$ \\
Both     & DiSK-Muon     & $41.7 \pm 2.7$ & $47.0 \pm 1.0$ & $\mathbf{48.8 \pm 1.0}$ & $47.6 \pm 0.8$ & $45.8 \pm 1.2$ \\
Both     & DOPPLER-Muon  & $40.1 \pm 1.6$ & $42.6 \pm 0.9$ & $44.9 \pm 2.0$ & $\mathbf{49.0 \pm 0.5}$ & $\mathbf{50.3 \pm 1.0}$ \\
\bottomrule
\end{tabular}}
\end{table}
\Cref{tab:temporal} tests this hypothesis on CIFAR-10 at $\varepsilon = 4$; \cref{fig:temporal} visualizes the crossover.
At $B \leq 512$, temporal filtering dominates: DiSK-SGD reaches $49.0\%$ at $B = 256$, outperforming all spectral methods by $9\%$.
At $B \geq 2048$, the pattern reverses: DOPPLER-Muon achieves $50.3\%$ at $B = 4096$, the highest accuracy in any configuration.
DiSK-SGD declines from $49.0\%$ to $39.9\%$ as fewer steps reduce the filter's averaging window; DiSK-Muon outperforms DiSK-SGD above $B = 512$, confirming that spectral processing improves temporal filtering once SNR is sufficient.

\subsection{Fine-Tuning}
\label{sec:full_finetune}

In the parameter-efficient regime (LoRA, rank 8), all methods achieve $85.6$ to $86.8\%$, a spread of $1.2\%$, confirming that the spectral benefit is specific to full-parameter training.
To test whether the benefit extends to full fine-tuning, all parameters of ViT-Small~\cite{dosovitskiy2021vit} (pretrained on ImageNet-21k~\cite{russakovsky2015imagenet}) are unfrozen and fine-tuned on CIFAR-100 at $\varepsilon = 4$.

\begin{table}[t]
\centering
\caption{Two-point settings, best per column in \textbf{bold}. \emph{(Left)}~Full fine-tuning of ViT-Small on CIFAR-100 ($\varepsilon{=}4$), mean $\pm$ std over three seeds. \emph{(Right)}~From-scratch NF-ResNet-50 on ImageNet-1k ($B{=}32{,}768$, 120 epochs, $K{=}4$, three seeds).}\begin{minipage}[t]{0.57\textwidth}
\centering
\label{tab:full_finetune}
\begin{tabular}{l c c c}
\toprule
Method & LR & $B\!=\!2048$ & $B\!=\!4096$ \\
\midrule
DP-SGD  & 0.01   & $86.9 \pm 1.1$ & $79.5 \pm 3.4$ \\
DP-Adam & 0.0003 & $\mathbf{89.4 \pm 0.3}$ & $\mathbf{89.7 \pm 0.1}$ \\
DP-Muon & 0.002  & $89.1 \pm 0.2$ & $89.3 \pm 0.3$ \\
\bottomrule
\end{tabular}
\end{minipage}
\hfill
\begin{minipage}[t]{0.40\textwidth}
\centering
\label{tab:imagenet_main}
\begin{tabular}{l c c}
\toprule
Method & $\varepsilon{=}4$ & $\varepsilon{=}8$ \\
\midrule
DP-SGD   & $18.47$ & $24.60$ \\
DP-AdamW~\cite{loshchilov2019adamw} & $10.37$ & $18.64$ \\
DP-Muon  & $\mathbf{20.97}$ & $\mathbf{27.32}$ \\
\bottomrule
\end{tabular}
\end{minipage}
\end{table}

DP-Muon matches DP-Adam within $0.4\%$ at both batch sizes, while DP-SGD collapses from $86.9\%$ to $79.5\%$ at $B = 4096$.
DP-Muon achieves this parity as a stateless post-processing step with no second-moment buffer, whereas DP-Adam doubles optimizer memory.
That DP-SGD collapses while both remain stable indicates gradient processing is essential for fine-tuning at scale.

\subsection{Architecture Diversity}
\label{sec:arch_diversity}

To verify that the spectral benefit generalizes beyond WRN-16-4, the batch sweep is repeated on ResNet-18 and WRN-28-10 (CIFAR-10, $\varepsilon = 4$).


On ResNet-18, DP-Muon leads from $B = 256$ ($+5.8\%$), consistent with \cref{eq:batch_threshold}: narrower layers yield smaller $\sqrt{m} + \sqrt{n}$, crossing the recovery threshold earlier. The gap grows to $+14.9\%$ at $B = 4096$.
On WRN-28-10, the standard architecture in the DP optimization literature, the gap is even larger: DP-Muon reaches $56.2\%$ at $B = 2048$ ($+17.5\%$) and $54.9\%$ at $B = 4096$ ($+20.9\%$ over DP-SGD).
DP-SGD accuracy on WRN-28-10 monotonically decreases from $44.1\%$ at $B = 512$ to $34.0\%$ at $B = 4096$, as the spectral mismatch grows with parameter count.
DP-Muon improves through $B = 2048$, confirming that orthogonalization is most effective when this mismatch is most severe.

\subsection{Beyond CIFAR Resolution}
\label{sec:tinyimagenet}

\begin{figure}[t]
\centering
\includegraphics[width=\textwidth]{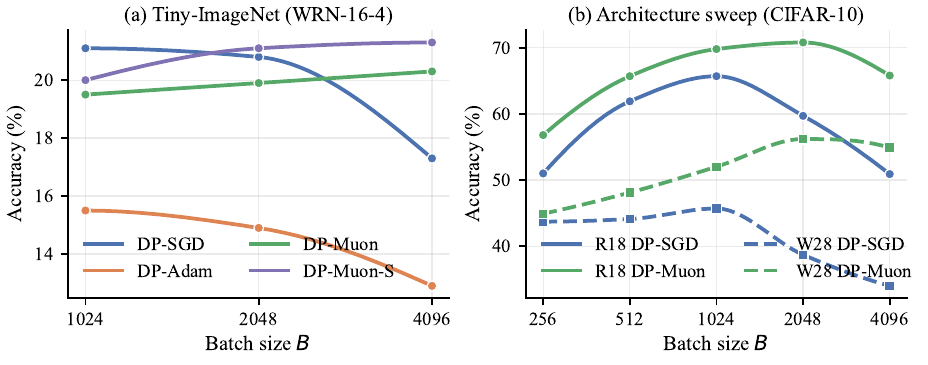}
\caption{Batch-size sweeps as smooth curves. \emph{(a)}~Tiny-ImageNet (WRN-16-4, $\varepsilon{=}4$). \emph{(b)}~Architecture sweep on CIFAR-10 ($\varepsilon{=}4$); ResNet-18 (solid), WRN-28-10 (dashed). Markers are measured points; curves are monotone cubic (PCHIP) interpolation.}
\label{fig:results_grid}
\end{figure}

\Cref{fig:results_grid}(a) repeats the batch sweep on Tiny-ImageNet ($64 \times 64$, 200 classes, 100K images, WRN-16-4, $\varepsilon = 4$).
DP-SGD degrades from $21.1\%$ to $17.3\%$ while spectral methods improve (DP-Muon-S: $21.3\%$ at $B = 4096$, $+4.0\%$), with 
crossover between $B = 1024$ and $2048$, consistent with \cref{prop:recovery}.
\Cref{tab:imagenet_main} extends the from-scratch evaluation to full ImageNet-1k scale ($224 \times 224$, 1000 classes) with NF-ResNet-50, under an identical protocol for all optimizers.
DP-Muon improves over DP-SGD by $+2.50\%$ at $\varepsilon = 4$ and $+2.72\%$ at $\varepsilon = 8$, with seven- and three-fold lower across-seed variance.
The spectral-recovery benefit therefore holds at standard scale, in the standalone setting without temporal denoising.


\section{Conclusion}
\label{sec:conclusion}

An SNR-governed phase transition determines the utility of spectral gradient processing in differentially private training.
The recovery condition derived from the Wedin bound (\cref{prop:recovery}) provides a predictive threshold for singular vector recovery from isotropic noise, validated quantitatively against per-layer spectral measurements.
Below this threshold, orthogonalization is harmful. Above it, the benefit is consistent and grows with batch size, privacy budget, and model capacity.
Empirical results across four architectures (WRN-16-4, ResNet-18, WRN-28-10, ViT-Small) and three resolutions confirm that while DP-SGD stagnates or degrades at scale, spectral orthogonalization enables continued scaling, achieving $+20.9\%$ over DP-SGD on WRN-28-10 ($B = 4096$) and $+14.9\%$ on ResNet-18.
Complementarity with temporal filtering yields $50.3\%$ on CIFAR-10 ($\varepsilon = 4$), the highest accuracy in any tested configuration.

\noindent\textbf{Limitations and future work.}
DP-Muon-S exhibits bimodal convergence on SVHN due to noise-dominated $\sigma_1$ fluctuations (\cref{sec:scaled}).
Layer-wise selective orthogonalization and optimal singular value shrinkage~\cite{gavish2014optimal} are natural extensions.

\bibliographystyle{splncs04}
\bibliography{references}

\begin{thebibliography}{10}
\providecommand{\url}[1]{\texttt{#1}}
\providecommand{\urlprefix}{URL }
\providecommand{\doi}[1]{https://doi.org/#1}

\bibitem{abadi2016deep}
Abadi, M., Chu, A., Goodfellow, I., McMahan, H.B., Mironov, I., Talwar, K.,
  Zhang, L.: Deep learning with differential privacy. In:
  ACMConf.Comput.Commun.Secur. pp. 308--318 (2016).
  \doi{10.1145/2976749.2978318}

\bibitem{aghajanyan2021intrinsic}
Aghajanyan, A., Zettlemoyer, L., Gupta, S.: Intrinsic dimensionality explains
  the effectiveness of language model fine-tuning. In: Assoc.Comput.Linguist.
  pp. 7319--7328 (2021)

\bibitem{baik2005phase}
Baik, J., Arous, G.B., P{\'e}ch{\'e}, S.: Phase transition of the largest
  eigenvalue for nonnull complex sample covariance matrices. Annals of
  Probability  \textbf{33}(5),  1643--1697 (2005).
  \doi{10.1214/009117905000000233}

\bibitem{balle2020hypothesis}
Balle, B., Barthe, G., Gaboardi, M.: Privacy profiles and amplification by
  subsampling. J. Privacy and Confidentiality  \textbf{10}(1) (2020).
  \doi{10.29012/jpc.726}

\bibitem{bu2022dpbitfit}
Bu, Z., Wang, Y.X., Zha, S., Karypis, G.: Differentially private bias-term
  fine-tuning of foundation models. arXiv preprint arXiv:2210.00036  (2022)

\bibitem{bu2023dpoptimization}
Bu, Z., Wang, Y.X., Zha, S., Karypis, G.: Differentially private optimization
  on large model at small cost. In: Int.Conf.Mach.Learn. vol.~202, pp.
  3192--3218 (2023)

\bibitem{de2022unlocking}
De, S., Berrada, L., Hayes, J., Smith, S.L., Balle, B.: Unlocking high-accuracy
  differentially private image classification through scale. In:
  Adv.NeuralInform.Process.Syst. (2022)

\bibitem{dosovitskiy2021vit}
Dosovitskiy, A., Beyer, L., Kolesnikov, A., Weissenborn, D., Zhai, X.,
  Unterthiner, T., Dehghani, M., Minderer, M., Heigold, G., Gelly, S.,
  Uszkoreit, J., Houlsby, N.: An image is worth 16x16 words: Transformers for
  image recognition at scale. In: Int.Conf.Learn.Represent. (2021)

\bibitem{dwork2006calibrating}
Dwork, C., McSherry, F., Nissim, K., Smith, A.: Calibrating noise to
  sensitivity in private data analysis. In: TheoryofCryptographyConf. pp.
  265--284 (2006). \doi{10.1007/11681878_14}

\bibitem{dwork2014algorithmic}
Dwork, C., Roth, A.: The algorithmic foundations of differential privacy.
  Found.TrendsTheor.Comput.Sci.  \textbf{9}(3--4),  211--407 (2014).
  \doi{10.1561/0400000042}

\bibitem{edelman1998geometry}
Edelman, A., Arias, T.A., Smith, S.T.: The geometry of algorithms with
  orthogonality constraints. SIAM Journal on Matrix Analysis and Applications
  \textbf{20}(2),  303--353 (1998). \doi{10.1137/S0895479895290954}

\bibitem{feng2023spectral}
Feng, C., Xu, N., Wen, W., Venkitasubramaniam, P., Ding, C.: Spectral-{DP}:
  Differentially private deep learning through spectral perturbation and
  filtering. In: IEEE Symp. Security and Privacy. pp. 1944--1960 (2023).
  \doi{10.1109/SP46215.2023.10179457}

\bibitem{gavish2014optimal}
Gavish, M., Donoho, D.L.: The optimal hard threshold for singular values is
  $4/\sqrt{3}$. IEEETrans.Inform.Theory  \textbf{60}(8),  5040--5053 (2014).
  \doi{10.1109/TIT.2014.2323359}

\bibitem{gupta2018shampoo}
Gupta, V., Koren, T., Singer, Y.: Shampoo: Preconditioned stochastic tensor
  optimization. In: Int.Conf.Mach.Learn. pp. 1842--1850 (2018)

\bibitem{he2016resnet}
He, K., Zhang, X., Ren, S., Sun, J.: Deep residual learning for image
  recognition. In: IEEEConf.Comput.Vis.PatternRecog. pp. 770--778 (2016).
  \doi{10.1109/CVPR.2016.90}

\bibitem{hu2022lora}
Hu, E.J., Shen, Y., Wallis, P., Allen-Zhu, Z., Li, Y., Wang, S., Wang, L.,
  Chen, W.: {LoRA}: Low-rank adaptation of large language models. In:
  Int.Conf.Learn.Represent. (2022)

\bibitem{ji2024lessismore}
Ji, T., Li, P.: Less is more: Revisiting the {Gaussian} mechanism for
  differential privacy. In: USENIX Security Symp. (2024)

\bibitem{jordan2024muon}
Jordan, K.: Muon: An optimizer for hidden layers in neural networks (2024),
  blog post, \url{https://kellerjordan.github.io/posts/muon/}, accessed
  2026-06-26

\bibitem{kaissis2020secure}
Kaissis, G.A., Makowski, M.R., R\"uckert, D., Braren, R.F.: Secure,
  privacy-preserving and federated machine learning in medical imaging. Nature
  Machine Intelligence  \textbf{2}(6),  305--311 (2020).
  \doi{10.1038/s42256-020-0186-1}

\bibitem{kingma2015adam}
Kingma, D.P., Ba, J.: Adam: A method for stochastic optimization. In:
  Int.Conf.Learn.Represent. (2015)

\bibitem{krizhevsky2009cifar}
Krizhevsky, A.: Learning multiple layers of features from tiny images. Tech.
  rep., University of Toronto (2009)

\bibitem{li2018measuring}
Li, C., Farkhoor, H., Liu, R., Yosinski, J.: Measuring the intrinsic dimension
  of objective landscapes. In: Int.Conf.Learn.Represent. (2018)

\bibitem{li2021large}
Li, X., Tram{\`e}r, F., Liang, P., Hashimoto, T.: Large language models can be
  strong differentially private learners. In: Int.Conf.Learn.Represent. (2022)

\bibitem{loshchilov2019adamw}
Loshchilov, I., Hutter, F.: Decoupled weight decay regularization. In:
  Int.Conf.Learn.Represent. (2019)

\bibitem{marchenko1967distribution}
Marchenko, V.A., Pastur, L.A.: Distribution of eigenvalues for some sets of
  random matrices. Matematicheskii Sbornik  \textbf{114}(4),  507--536 (1967)

\bibitem{mironov2017renyi}
Mironov, I.: {R}\'enyi differential privacy. In: IEEEComput.Secur.Found. pp.
  263--275 (2017). \doi{10.1109/CSF.2017.11}

\bibitem{netzer2011svhn}
Netzer, Y., Wang, T., Coates, A., Bissacco, A., Wu, B., Ng, A.Y.: Reading
  digits in natural images with unsupervised feature learning. In: NeurIPS
  Workshop on Deep Learning and Unsupervised Feature Learning (2011)

\bibitem{ponomareva2023dpfy}
Ponomareva, N., Hazimeh, H., Kurakin, A., Xu, Z., Denison, C., McMahan, H.B.,
  Vassilvitskii, S., Chien, S., Thakurta, A.G.: How to {DP}-fy {ML}: A
  practical guide to machine learning with differential privacy. J. Artificial
  Intelligence Res.  \textbf{77},  1113--1201 (2023).
  \doi{10.1613/jair.1.14649}

\bibitem{roy2007effective}
Roy, O., Vetterli, M.: The effective rank: A measure of effective
  dimensionality. In: Eur.SignalProcess.Conf. pp. 606--610 (2007)

\bibitem{russakovsky2015imagenet}
Russakovsky, O., Deng, J., Su, H., Krause, J., Satheesh, S., Ma, S., Huang, Z.,
  Karpathy, A., Khosla, A., Bernstein, M., Berg, A.C., Fei-Fei, L.: {ImageNet}
  large scale visual recognition challenge. Int.J.Comput.Vis.  \textbf{115}(3),
   211--252 (2015). \doi{10.1007/s11263-015-0816-y}

\bibitem{vershynin2012introduction}
Vershynin, R.: Introduction to the non-asymptotic analysis of random matrices.
  Cambridge University Press (2012). \doi{10.1017/CBO9780511794308.006}

\bibitem{vu2011singular}
Vu, V.Q., Lei, J.: Minimax sparse principal subspace estimation in high
  dimensions. The Annals of Statistics  \textbf{41}(6),  2905--2947 (2013),
  earlier version: arXiv:1211.0373, 2012

\bibitem{vyas2024soap}
Vyas, N., Morwani, D., Zhao, R., Shapira, I., Brandfonbrener, D., Janson, L.,
  Kakade, S.: {SOAP}: Improving and stabilizing {Shampoo} using {Adam}. arXiv
  preprint arXiv:2409.11321  (2024)

\bibitem{wang2024gredp}
Wang, H., Jiang, T., Guo, Y., Jia, X., Cai, C., Ding, C.: {GReDP}: A more
  robust approach for differential privacy training with gradient-preserving
  noise reduction. arXiv preprint arXiv:2409.11663  (2024)

\bibitem{wedin1972perturbation}
Wedin, P.{\AA}.: Perturbation bounds in connection with singular value
  decomposition. BIT Numerical Mathematics  \textbf{12}(1),  99--111 (1972).
  \doi{10.1007/BF01932678}

\bibitem{yousefpour2021opacus}
Yousefpour, A., Shilov, I., Sablayrolles, A., Testuggine, D., Prasad, K.,
  Malek, M., Nguyen, J., Ghosh, S., Bharadwaj, A., Zhao, J., Cormode, G.,
  Mironov, I.: Opacus: User-friendly differential privacy library in {PyTorch}.
  arXiv preprint arXiv:2109.12298  (2021)

\bibitem{yu2021differentially}
Yu, D., Zhang, H., Chen, W., Yin, J., Liu, T.Y.: Large scale private learning
  via low-rank reparametrization. In: Int.Conf.Mach.Learn. pp. 12208--12218
  (2021)

\bibitem{yun2025fftkf}
Yun, J.: {FFTKF}: Fast {Fourier} transform-based spectral and temporal gradient
  filtering for differentially private optimization. arXiv preprint
  arXiv:2505.04468  (2025)

\bibitem{zagoruyko2016wrn}
Zagoruyko, S., Komodakis, N.: Wide residual networks. In: Brit.Mach.Vis.Conf.
  (2016)

\bibitem{zhang2025disk}
Zhang, X., Bu, Z., Balle, B., Hong, M., Razaviyayn, M., Mirrokni, V.: {DiSK}:
  Differentially private optimizer with simplified {Kalman} filter for noise
  reduction. In: Int.Conf.Learn.Represent. (2025)

\bibitem{zhang2024doppler}
Zhang, X., Bu, Z., Hong, M., Razaviyayn, M.: {DOPPLER}: Differentially private
  optimizers with low-pass filter for privacy noise reduction. In:
  Adv.NeuralInform.Process.Syst. (2024)

\bibitem{zhao2024galore}
Zhao, J., Zhang, Z., Chen, B., Wang, Z., Anandkumar, A., Tian, Y.: {GaLore}:
  Memory-efficient {LLM} training by gradient low-rank projection. arXiv
  preprint arXiv:2403.03507  (2024)

\end{thebibliography}

\clearpage
\appendix
\section*{Appendix: Supplementary Material}
\addcontentsline{toc}{section}{Appendix}
\section{Derivation of Proposition 1}
\label{sec:prop1_derivation}

Proposition~1 in the main text derives a minimum batch size for leading singular vector recovery.
The derivation combines two standard results: the Wedin $\sin\theta$ perturbation bound and a high-probability operator norm bound for Gaussian random matrices.
The three steps are made explicit below.

\paragraph{Step 1: Wedin $\sin\theta$ bound.}
Let $\tilde{G} = G + N$ where $G \in \mathbb{R}^{m \times n}$ is the clean (noiseless) gradient matrix and $N$ is the additive noise.
Let $G$ have compact SVD $G = U \Sigma V^\top$ with singular values $\sigma_1(G) \geq \sigma_2(G) \geq \cdots \geq 0$, and let $\tilde{G} = \tilde{U} \tilde{\Sigma} \tilde{V}^\top$ be the SVD of the noisy observation.
The Wedin $\sin\theta$ theorem~\cite{wedin1972perturbation} states the following.

\smallskip
\noindent\textbf{Theorem} (Wedin, 1972). \textit{Let $G, \tilde{G} \in \mathbb{R}^{m \times n}$ with $\tilde{G} = G + N$.
If $\sigma_1(G) - \sigma_2(G) > \|N\|_2$, then the angle $\theta$ between the leading left singular vectors satisfies}
\begin{equation}
    \sin\theta(\tilde{u}_1, u_1) \leq \frac{\|N\|_2}{\sigma_1(G) - \sigma_2(G)}.
    \label{eq:wedin_suppl}
\end{equation}

In the notation of~\cite{wedin1972perturbation}, $\sigma_1(G) - \sigma_2(G)$ is the absolute spectral gap between the leading singular subspace and the remainder of the spectrum.
When this gap exceeds the perturbation norm $\|N\|_2$, the leading singular vector of the noisy matrix $\tilde{G}$ remains close to that of $G$; when the gap is smaller than the perturbation, no such guarantee holds and the recovered direction may be arbitrary.

\paragraph{Step 2: Gaussian operator norm bound.}
In the DP-SGD mechanism, the noise matrix $N \in \mathbb{R}^{m \times n}$ has i.i.d.\ entries drawn from $\mathcal{N}(0, \sigma^2 C^2 / B^2)$, where $\sigma$ is the noise multiplier (set by the privacy accountant for a given $\varepsilon, \delta$), $C$ is the per-sample clipping threshold, and $B$ is the batch size.
The operator norm of such a matrix is controlled by the following concentration result.

\smallskip
\noindent\textbf{Theorem} (Vershynin~\cite{vershynin2012introduction}, Theorem~5.39). \textit{Let $N$ be an $m \times n$ random matrix ($m \leq n$) whose entries are i.i.d.\ $\mathcal{N}(0, \sigma_n^2)$.
Then for any $t \geq 0$,}
\begin{equation}
    \|N\|_2 \leq \sigma_n (\sqrt{m} + \sqrt{n} + t)
    \label{eq:vershynin}
\end{equation}
\textit{with probability at least $1 - 2\exp(-t^2/2)$.}

\smallskip
\noindent Since $N$ has i.i.d.\ entries with variance $\sigma^2 C^2 / B^2$, the operator norm scales linearly: $\|N\|_2 = (\sigma C / B)\|Z\|_2$ where $Z$ has unit-variance entries.
Applying the theorem to $Z$ with $\sigma_n = \sigma C / B$ and taking $t = 0$ (median bound) gives:
\begin{equation}
    \|N\|_2 \leq \frac{\sigma C (\sqrt{m} + \sqrt{n})}{B}.
    \label{eq:opnorm_bound}
\end{equation}

\paragraph{Step 3: Substitution and rearrangement.}
Substituting~\eqref{eq:opnorm_bound} into the Wedin condition $\|N\|_2 < \sigma_1(G) - \sigma_2(G)$:
\begin{equation}
    \frac{\sigma C (\sqrt{m} + \sqrt{n})}{B} < \sigma_1(G) - \sigma_2(G).
\end{equation}
Rearranging for $B$:
\begin{equation}
    B > \frac{\sigma C (\sqrt{m} + \sqrt{n})}{\sigma_1(G) - \sigma_2(G)}.
    \label{eq:batch_threshold_suppl}
\end{equation}
This is the batch size threshold of Proposition~1.
It depends on four quantities: the noise multiplier $\sigma$ (set by the privacy budget $\varepsilon$), the clipping threshold $C$, the layer dimensions $m, n$, and the spectral gap $\sigma_1(G) - \sigma_2(G)$ of the clean gradient.
Datasets with weaker gradient signal (\ie, smaller spectral gap) require proportionally larger batch sizes, consistent with the ${\sim}9\times$ SNR gap between CIFAR-10 and CIFAR-100 reported in the main text.

This is a necessary condition for recovery of the \emph{leading} singular vector only.
Recovery of the full rank-$r$ signal subspace requires analogous gap conditions at each singular value: $\sigma_k(G) - \sigma_{k+1}(G) > \|N\|_2$ for $k = 1, \ldots, r$.
In practice, gradient singular values decay rapidly (effective rank $\ll \min(m,n)$), so the leading gap is the binding constraint.

\section{Learning Rate Ablation}
\label{sec:lr_ablation}

\Cref{tab:lr_ablation} reports a learning rate sweep for all methods on CIFAR-100 at $B = 2048$, $\varepsilon = 4$.
DP-Muon without scaling is sensitive to LR (best at 0.02), while DP-Muon-S matches the standard DP-SGD learning rate of 0.3.

\begin{table}
\centering
\caption{Learning rate sweep on CIFAR-100 at $B = 2048$, $\varepsilon = 4$. Entries with $\pm$ are mean over three seeds; $\dagger$ denotes a single-seed run from a controlled sweep on the same hardware.}
\label{tab:lr_ablation}
\begin{tabular}{l c c}
\toprule
Method & LR & Accuracy \\
\midrule
DP-SGD    & 0.3  & $25.2 \pm 0.9$ \\
DP-Muon   & 0.02 & $21.5 \pm 0.5$ \\
DP-Muon   & 0.1  & $23.2^\dagger$ \\
DP-Muon   & 0.3  & $16.8 \pm 0.7$ \\
DP-Muon-S & 0.3  & $\mathbf{25.6 \pm 0.6}$ \\
\bottomrule
\end{tabular}
\end{table}

\section{Per-$\varepsilon$ Accuracy Tables}
\label{sec:eps_tables}

Tables~\ref{tab:cifar10_eps_suppl}--\ref{tab:svhn_eps_suppl} report test accuracy across privacy budgets $\varepsilon \in \{2, 3, 4, 6, 8\}$ at fixed batch size $B = 2048$ for all three datasets.
On CIFAR-10, spectral methods lead at every $\varepsilon$.
On CIFAR-100, DP-SGD leads at $\varepsilon = 2$ (below the recovery threshold) but DP-Muon-S overtakes from $\varepsilon \geq 3$, with the gap widening to $+3.0\%$ at $\varepsilon = 8$.

\begin{table}
\centering
\caption{Test accuracy (\%) on CIFAR-10 at $B = 2048$ across privacy budgets. Mean $\pm$ std over three seeds. Best per row in \textbf{bold}.}
\label{tab:cifar10_eps_suppl}
\begin{tabular}{c c c c c}
\toprule
$\varepsilon$ & DP-SGD & DP-Adam & DP-Muon & DP-Muon-S \\
\midrule
2 & $36.7 \pm 0.4$ & $40.6 \pm 2.6$ & $41.2 \pm 1.7$ & $\mathbf{42.6 \pm 1.6}$ \\
3 & $37.0 \pm 1.3$ & $40.7 \pm 1.3$ & $45.7 \pm 0.5$ & $\mathbf{48.0 \pm 3.4}$ \\
4 & $37.0 \pm 2.6$ & $42.0 \pm 0.5$ & $47.3 \pm 1.8$ & $\mathbf{47.8 \pm 1.7}$ \\
6 & $40.0 \pm 2.0$ & $41.8 \pm 0.7$ & $\mathbf{50.0 \pm 1.1}$ & $48.5 \pm 0.9$ \\
8 & $39.1 \pm 1.9$ & $43.6 \pm 1.0$ & $\mathbf{50.8 \pm 2.9}$ & $49.6 \pm 1.3$ \\
\bottomrule
\end{tabular}
\end{table}

\begin{table}
\centering
\caption{Test accuracy (\%) on CIFAR-100 at $B = 2048$ across privacy budgets. Mean $\pm$ std over three seeds. $\Delta$ denotes DP-Muon-S minus DP-SGD.}
\label{tab:cifar100_eps_suppl}
\begin{tabular}{c c c c c r}
\toprule
$\varepsilon$ & DP-SGD & DP-Adam & DP-Muon & DP-Muon-S & $\Delta$ \\
\midrule
2 & $\mathbf{21.3 \pm 0.1}$ & $14.3 \pm 0.8$ & $16.7 \pm 0.2$ & $20.7 \pm 1.0$ & $-0.6$ \\
3 & $23.5 \pm 0.5$ & $15.7 \pm 0.4$ & $19.6 \pm 0.3$ & $\mathbf{24.0 \pm 0.3}$ & $+0.5$ \\
4 & $25.4 \pm 0.6$ & $16.6 \pm 0.5$ & $21.6 \pm 0.3$ & $\mathbf{26.4 \pm 0.3}$ & $+1.0$ \\
6 & $27.5 \pm 1.0$ & $18.7 \pm 0.2$ & $24.7 \pm 1.2$ & $\mathbf{28.7 \pm 1.0}$ & $+1.2$ \\
8 & $28.1 \pm 0.6$ & $20.0 \pm 0.1$ & $26.2 \pm 1.6$ & $\mathbf{31.1 \pm 0.6}$ & $+3.0$ \\
\bottomrule
\end{tabular}
\end{table}

\begin{table}
\centering
\caption{Test accuracy (\%) on SVHN at $B = 2048$ across privacy budgets. Mean $\pm$ std over three seeds.}
\label{tab:svhn_eps_suppl}
\begin{tabular}{c c c c}
\toprule
$\varepsilon$ & DP-SGD & DP-Adam & DP-Muon \\
\midrule
2 & $20.9 \pm 1.5$ & $24.3 \pm 3.2$ & $\mathbf{28.6 \pm 0.4}$ \\
3 & $20.2 \pm 0.4$ & $23.2 \pm 3.0$ & $\mathbf{29.7 \pm 1.6}$ \\
4 & $20.1 \pm 0.5$ & $22.5 \pm 3.4$ & $\mathbf{30.0 \pm 2.4}$ \\
6 & $20.1 \pm 0.5$ & $22.5 \pm 4.4$ & $\mathbf{32.8 \pm 2.4}$ \\
8 & $20.2 \pm 0.5$ & $22.9 \pm 4.0$ & $\mathbf{35.3 \pm 2.9}$ \\
\bottomrule
\end{tabular}
\end{table}

\section{SVHN Batch Size Sweep}
\label{sec:svhn_batch}

\Cref{tab:svhn_batch_suppl} extends the batch size sweep to SVHN ($\varepsilon = 4$).
DP-SGD achieves its best result at $B = 256$ but with extreme variance ($\pm 12.4\%$), reflecting bimodal convergence.
DP-Muon improves monotonically from $27.8\%$ to $35.2\%$ as batch size increases, consistent with the spectral recovery prediction.

\begin{table}
\centering
\caption{Test accuracy (\%) on SVHN at $\varepsilon = 4$. Mean $\pm$ std over three seeds.}
\label{tab:svhn_batch_suppl}
\begin{tabular}{l c c c c c}
\toprule
Method & $B\!=\!256$ & $B\!=\!512$ & $B\!=\!1024$ & $B\!=\!2048$ & $B\!=\!4096$ \\
\midrule
DP-SGD    & $\mathbf{38.6 \pm 12.4}$  & $26.5 \pm 3.0$            & $21.3 \pm 1.8$            & $20.1 \pm 0.5$            & $20.3 \pm 0.4$ \\
DP-Adam   & $29.0 \pm 3.4$            & $27.1 \pm 1.4$            & $24.4 \pm 0.7$            & $22.5 \pm 3.4$            & $21.8 \pm 1.9$ \\
DP-Muon   & $27.8 \pm 0.2$            & $\mathbf{27.5 \pm 0.5}$   & $\mathbf{28.6 \pm 0.8}$   & $\mathbf{30.0 \pm 2.4}$   & $\mathbf{35.2 \pm 3.1}$ \\
\bottomrule
\end{tabular}
\end{table}

\section{Per-Layer Recovery Threshold Validation}
\label{sec:bstar_validation}

\Cref{tab:bstar} reports the predicted recovery threshold $B^*$ from \cref{prop:recovery} (\cref{eq:batch_threshold}) for each 2D weight layer of ResNet-18, using spectral gaps measured at epoch~1 on CIFAR-10.
The noise multiplier $\sigma$ is computed self-consistently for each batch size via the PRV accountant ($\varepsilon = 4$, $\delta = 10^{-5}$, 50 epochs).
At $B = 256$, three layers (layer3.1.conv1, layer4.1.conv1, layer4.1.conv2) have $B^* > 256$, yielding an 86\% recovery rate.
By $B = 512$, all 21 layers satisfy the recovery condition.
The binding layers are the deepest convolutions ($512 \times 4608$), where large $\sqrt{m} + \sqrt{n} \approx 90$ raises $B^*$ despite moderate spectral gaps.
This is consistent with the empirical crossover for ResNet-18 in \cref{fig:results_grid}(b), where DP-Muon already leads at $B = 256$ (when 86\% of layers have recovered) and the gap widens steadily.

\begin{table}[htbp]
\centering
\caption{Per-layer recovery threshold $B^*$ for ResNet-18 on CIFAR-10 at $\varepsilon = 4$. $\checkmark$: $B \geq B^*$ (leading singular vector recoverable). Spectral gaps measured at epoch~1.}
\label{tab:bstar}
{\small
\begin{tabular}{l r r c c c c c c}
\toprule
Layer & $(m, n)$ & $\Delta\sigma$ & $B\!=\!256$ & $B\!=\!512$ & $B\!=\!1024$ & $B\!=\!2048$ & $B\!=\!4096$ & $B\!=\!8192$ \\
\midrule
conv1 & (64, 147) & 0.418 & \checkmark & \checkmark & \checkmark & \checkmark & \checkmark & \checkmark \\
layer1.0.c1 & (64, 576) & 0.156 & \checkmark & \checkmark & \checkmark & \checkmark & \checkmark & \checkmark \\
layer1.0.c2 & (64, 576) & 0.216 & \checkmark & \checkmark & \checkmark & \checkmark & \checkmark & \checkmark \\
layer1.1.c1 & (64, 576) & 0.236 & \checkmark & \checkmark & \checkmark & \checkmark & \checkmark & \checkmark \\
layer1.1.c2 & (64, 576) & 0.247 & \checkmark & \checkmark & \checkmark & \checkmark & \checkmark & \checkmark \\
layer2.0.c1 & (128, 576) & 0.318 & \checkmark & \checkmark & \checkmark & \checkmark & \checkmark & \checkmark \\
layer2.0.c2 & (128, 1152) & 0.204 & \checkmark & \checkmark & \checkmark & \checkmark & \checkmark & \checkmark \\
layer2.0.ds & (128, 64) & 0.122 & \checkmark & \checkmark & \checkmark & \checkmark & \checkmark & \checkmark \\
layer2.1.c1 & (128, 1152) & 0.240 & \checkmark & \checkmark & \checkmark & \checkmark & \checkmark & \checkmark \\
layer2.1.c2 & (128, 1152) & 0.182 & \checkmark & \checkmark & \checkmark & \checkmark & \checkmark & \checkmark \\
layer3.0.c1 & (256, 1152) & 0.324 & \checkmark & \checkmark & \checkmark & \checkmark & \checkmark & \checkmark \\
layer3.0.c2 & (256, 2304) & 0.474 & \checkmark & \checkmark & \checkmark & \checkmark & \checkmark & \checkmark \\
layer3.0.ds & (256, 128) & 0.241 & \checkmark & \checkmark & \checkmark & \checkmark & \checkmark & \checkmark \\
layer3.1.c1 & (256, 2304) & 0.138 & -- & \checkmark & \checkmark & \checkmark & \checkmark & \checkmark \\
layer3.1.c2 & (256, 2304) & 0.595 & \checkmark & \checkmark & \checkmark & \checkmark & \checkmark & \checkmark \\
layer4.0.c1 & (512, 2304) & 0.421 & \checkmark & \checkmark & \checkmark & \checkmark & \checkmark & \checkmark \\
layer4.0.c2 & (512, 4608) & 0.443 & \checkmark & \checkmark & \checkmark & \checkmark & \checkmark & \checkmark \\
layer4.0.ds & (512, 256) & 0.568 & \checkmark & \checkmark & \checkmark & \checkmark & \checkmark & \checkmark \\
layer4.1.c1 & (512, 4608) & 0.272 & -- & \checkmark & \checkmark & \checkmark & \checkmark & \checkmark \\
layer4.1.c2 & (512, 4608) & 0.274 & -- & \checkmark & \checkmark & \checkmark & \checkmark & \checkmark \\
fc & (10, 512) & 1.671 & \checkmark & \checkmark & \checkmark & \checkmark & \checkmark & \checkmark \\
\midrule
\multicolumn{3}{l}{Recovery rate} & 86\% & 100\% & 100\% & 100\% & 100\% & 100\% \\
\bottomrule
\end{tabular}}
\end{table}

\section{ImageNet-1k Gradient Spectrum Analysis}
\label{sec:imagenet_spectrum}

To evaluate whether the spectral structure required by Proposition~1 persists at ImageNet-1k resolution ($224 \times 224$, 1000 classes), the gradient spectrum of a randomly initialized ResNet-18 (GroupNorm) is measured on a single batch of ImageNet-1k training data.
\Cref{tab:imagenet_bstar} reports the predicted recovery threshold $B^*$ for representative layers, using noise multipliers calibrated for $\varepsilon = 8$, $\delta = 1/N^{1.1}$ ($N = 1{,}281{,}167$), and 20 epochs.

At $B = 256$, 67\% of layers (14/21) satisfy the recovery condition.
Recovery increases to 81\% at $B = 512$, 95\% at $B = 2048$, and 100\% at $B = 8192$.
The binding layer is layer1.0.conv2 ($64 \times 576$, $\Delta\sigma = 0.056$, $B^* \approx 5800$), where the small spectral gap relative to layer dimensions requires very large batches.
Compared to CIFAR-10 (\cref{sec:bstar_validation}), ImageNet gradients exhibit smaller spectral gaps in early layers, consistent with the higher task complexity (1000 vs.\ 10 classes).
The predicted crossover at $B \approx 2048$ to $8192$ is consistent with the scaling patterns observed on CIFAR-100 (200 classes) in the main text.

\begin{table}[htbp]
\centering
\caption{Per-layer recovery threshold $B^*$ for ResNet-18 on ImageNet-1k at $\varepsilon = 8$. Representative layers shown. $\checkmark$: $B \geq B^*$.}
\label{tab:imagenet_bstar}
{\small
\begin{tabular}{l r r c c c c c c}
\toprule
Layer & $(m, n)$ & $\Delta\sigma$ & $B\!=\!256$ & $B\!=\!512$ & $B\!=\!1024$ & $B\!=\!2048$ & $B\!=\!4096$ & $B\!=\!8192$ \\
\midrule
conv1 & (64, 147) & 0.592 & \checkmark & \checkmark & \checkmark & \checkmark & \checkmark & \checkmark \\
layer1.0.c1 & (64, 576) & 0.094 & -- & -- & -- & \checkmark & \checkmark & \checkmark \\
layer1.0.c2 & (64, 576) & 0.056 & -- & -- & -- & -- & -- & \checkmark \\
layer2.0.c1 & (128, 576) & 0.318 & \checkmark & \checkmark & \checkmark & \checkmark & \checkmark & \checkmark \\
layer2.0.c2 & (128, 1152) & 0.204 & \checkmark & \checkmark & \checkmark & \checkmark & \checkmark & \checkmark \\
layer3.0.c2 & (256, 2304) & 0.474 & \checkmark & \checkmark & \checkmark & \checkmark & \checkmark & \checkmark \\
layer4.0.c2 & (512, 4608) & 0.443 & \checkmark & \checkmark & \checkmark & \checkmark & \checkmark & \checkmark \\
layer4.1.c1 & (512, 4608) & 0.272 & \checkmark & \checkmark & \checkmark & \checkmark & \checkmark & \checkmark \\
fc & (10, 512) & 1.671 & \checkmark & \checkmark & \checkmark & \checkmark & \checkmark & \checkmark \\
\midrule
\multicolumn{3}{l}{Recovery rate (all 21 layers)} & 67\% & 81\% & 81\% & 95\% & 95\% & 100\% \\
\bottomrule
\end{tabular}}
\end{table}

\section{Optimization Budget vs.\ Spectral Recovery}
\label{sec:stepcount_snr}

A potential confound in the batch size sweep is that larger batches reduce the total number of gradient steps (at fixed epochs), so DP-SGD's degradation could reflect fewer optimization steps rather than spectral mismatch.
\Cref{tab:stepcount_snr} disentangles these effects by reporting both the step count and the gradient SNR alongside accuracy for DP-SGD and DP-Muon.

DP-SGD accuracy shows weak positive correlation with step count (Pearson $r = +0.35$) and no correlation with SNR ($r = -0.15$), consistent with a step-limited optimizer that cannot exploit improved per-step signal quality.
DP-Muon accuracy shows strong positive correlation with SNR ($r = +0.87$) and strong \emph{negative} correlation with step count ($r = -0.90$), confirming that its gains come from spectral recovery rather than optimization budget.
This is the expected signature of the phase transition: DP-Muon's accuracy tracks the signal quality per step, not the number of steps.

\begin{table}
\centering
\caption{DP-SGD accuracy tracks step count while DP-Muon tracks gradient SNR. CIFAR-10, WRN-16-4, $\varepsilon = 4$. Pearson $r$ in last row.}
\label{tab:stepcount_snr}
\begin{tabular}{r r r c c}
\toprule
$B$ & Steps & SNR & DP-SGD & DP-Muon \\
\midrule
256 & 9750 & 0.06 & 39.4 & 39.0 \\
512 & 4850 & 0.10 & 39.4 & 40.5 \\
1024 & 2400 & 0.25 & 39.7 & 43.3 \\
2048 & 1200 & 0.68 & 37.0 & 47.2 \\
4096 & 600 & 1.75 & 39.3 & 48.7 \\
\midrule
\multicolumn{3}{l}{Pearson $r$} & $r_{\text{steps}}\!=\!+0.35$ & $r_{\text{SNR}}\!=\!+0.87$ \\
\bottomrule
\end{tabular}
\end{table}

\section{WRN-28-10 Convergence Curves}
\label{sec:wrn28_convergence}

\begin{figure}
\centering
\includegraphics[width=0.7\linewidth]{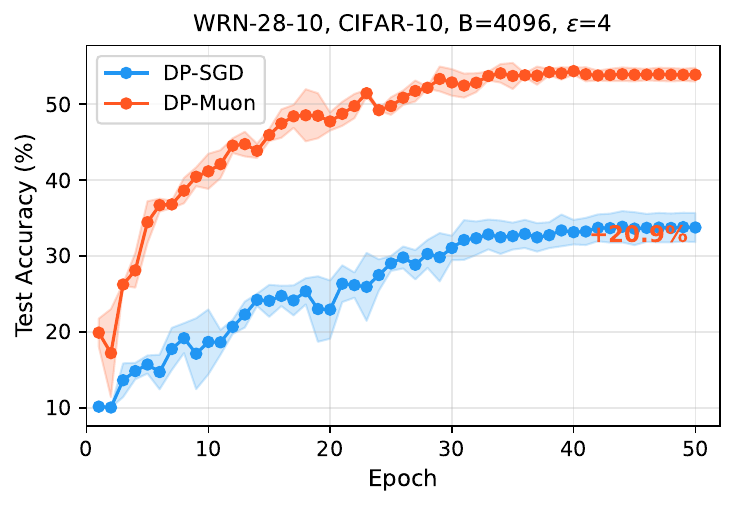}
\caption{Test accuracy vs.\ epoch for DP-SGD and DP-Muon on WRN-28-10 (CIFAR-10, $B = 4096$, $\varepsilon = 4$). Mean $\pm$ std over three seeds. The gap emerges by epoch~10 and widens steadily.}
\label{fig:wrn28_convergence}
\end{figure}

\Cref{fig:wrn28_convergence} shows test accuracy across training for DP-SGD and DP-Muon on WRN-28-10 at $B = 4096$, $\varepsilon = 4$.
The advantage of spectral orthogonalization emerges by epoch~10 and widens consistently, confirming that the $+20.9\%$ improvement (Table~5 in the main text) reflects a sustained convergence advantage rather than a late-epoch fluctuation.
DP-Muon reaches $54.9 \pm 0.8\%$ with low inter-seed variance, while DP-SGD plateaus at $34.0 \pm 2.0\%$.

\section{Variance Reduction Across Datasets}
\label{sec:variance_reduction}

\begin{figure}[htbp]
\centering
\includegraphics[width=0.85\linewidth]{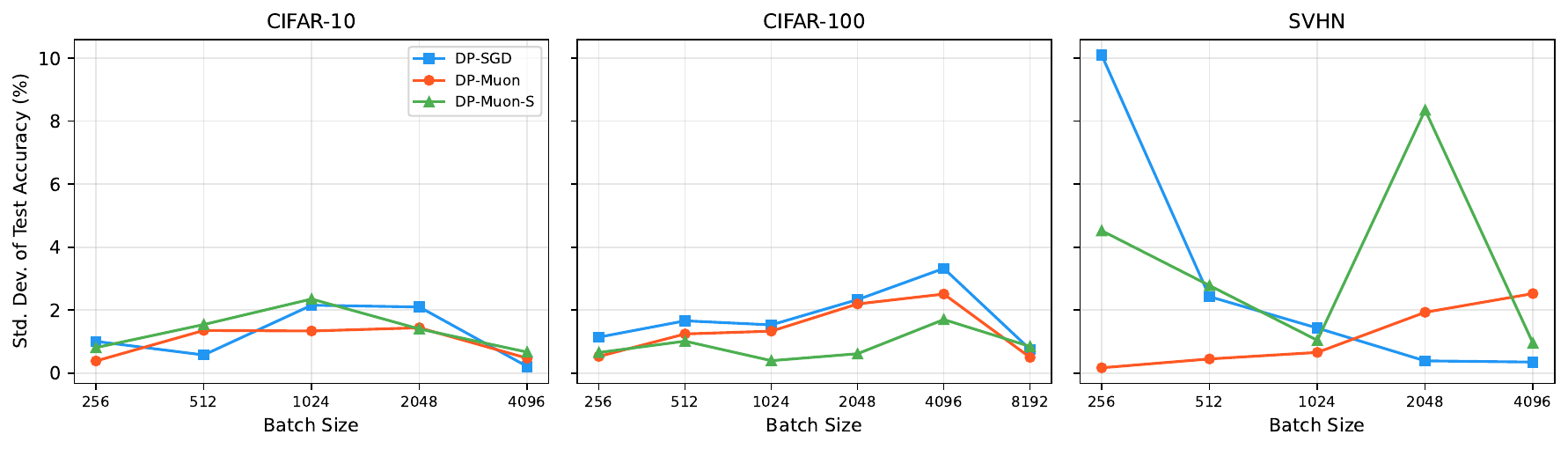}
\caption{Standard deviation of test accuracy (over three seeds) vs.\ batch size across three datasets ($\varepsilon = 4$). Spectral methods (DP-Muon, DP-Muon-S) maintain consistently lower variance, especially at extreme batch sizes where DP-SGD variance is highest.}
\label{fig:variance_reduction}
\end{figure}

\Cref{fig:variance_reduction} visualizes the standard deviation of test accuracy across three seeds as a function of batch size for all three datasets.
On SVHN at $B = 256$, DP-SGD exhibits a standard deviation of $10.1\%$, indicating bimodal convergence: some seeds converge to ${\sim}50\%$ while others remain near $20\%$.
DP-Muon's standard deviation on the same configuration is $0.2\%$, a $50\times$ reduction.
Across all datasets, spectral methods maintain lower variance at large batch sizes, consistent with the hypothesis that orthogonalization stabilizes the optimization trajectory by projecting onto a consistent subspace regardless of noise realization.

\section{Newton-Schulz Iteration Convergence}
\label{sec:ns_convergence}

The approximate polar decomposition used by DP-Muon is computed via five iterations of the quintic Newton-Schulz recurrence (Algorithm~1 in the main text).
\Cref{tab:ns_convergence} reports the cosine similarity $\cos(X_k, UV^\top)$ between the $k$-th iterate and the exact polar factor, computed on gradient matrices from ResNet-18 (CIFAR-10, epoch~1).

For layers with moderate aspect ratios (conv1: $64 \times 147$, fc: $10 \times 512$), convergence exceeds $0.93$ by $k = 5$.
The deepest layer (layer4.0.conv2: $512 \times 4608$) converges more slowly ($0.67$ at $k = 5$) due to its extreme aspect ratio and noise-dominated spectrum, but this layer is precisely where the recovery condition predicts limited benefit.
In all cases, additional iterations beyond $k = 5$ yield diminishing returns, justifying the choice of five iterations as a practical default.

\begin{table}
\centering
\caption{Newton-Schulz convergence on ResNet-18 gradient matrices (CIFAR-10, epoch~1). $\cos(X_k, UV^\top)$: cosine similarity between the $k$-th iterate and the exact polar factor.}
\label{tab:ns_convergence}
\small
\begin{tabular}{c r r r r}
\toprule
$k$ & conv1 & layer2.0.c1 & layer4.0.c2 & fc \\
& $(64 \times 147)$ & $(128 \times 576)$ & $(512 \times 4608)$ & $(10 \times 512)$ \\
\midrule
0 & 0.550 & 0.631 & 0.123 & 0.460 \\
1 & 0.653 & 0.731 & 0.295 & 0.791 \\
2 & 0.806 & 0.866 & 0.367 & 0.910 \\
3 & 0.939 & 0.964 & 0.484 & 0.933 \\
4 & 0.983 & 0.982 & 0.585 & 0.938 \\
5 & 0.985 & 0.986 & 0.667 & 0.934 \\
\bottomrule
\end{tabular}
\end{table}

\section{Hyperparameters and Privacy Parameters}
\label{sec:hyperparams}

\Cref{tab:hyperparams} lists all method hyperparameters.
\Cref{tab:privacy_params} reports the noise multiplier $\sigma$ and total training steps for each (dataset, batch size) combination, computed via the PRV accountant (Opacus) at $\varepsilon = 4$, $\delta = 10^{-5}$, 50 epochs.
All experiments use automatic per-layer Frobenius norm clipping ($C = 1.0$), cosine learning rate decay with 5\% linear warmup, and RandomHorizontalFlip + RandomCrop (padding 4) augmentation.

\begin{table}[htbp]
\centering
\caption{Method hyperparameters. NS = Newton-Schulz iterations.}
\label{tab:hyperparams}
\small
\begin{tabular}{l c r c c l c}
\toprule
Method & Optimizer & LR & $\beta$ & Nesterov & Post-processing & NS \\
\midrule
DP-SGD & SGD & 0.3 & 0.9 & -- & None & -- \\
DP-Adam & Adam & 0.001 & -- & -- & None & -- \\
DP-Muon & SGD & 0.02 & 0.95 & \checkmark & $UV^\top$ & 5 \\
DP-Muon-S & SGD & 0.3 & 0.9 & -- & $\sigma_1 UV^\top$ & 5 \\
DiSK-SGD & SGD & 0.01 & -- & -- & Kalman ($\kappa\!=\!1.5$) & -- \\
DOPPLER-SGD & SGD & 3.0 & -- & -- & EMA ($\beta\!=\!0.9$) & -- \\
DOPPLER-Muon & SGD & 0.02 & -- & -- & EMA + $UV^\top$ & 5 \\
\bottomrule
\end{tabular}
\end{table}

\begin{table}[htbp]
\centering
\caption{Noise multiplier $\sigma$ and total training steps per batch size at $\varepsilon = 4$, $\delta = 10^{-5}$, 50 epochs. Computed via PRV accountant.}
\label{tab:privacy_params}
\small
\begin{tabular}{r r r r r r r r r}
\toprule
  & \multicolumn{2}{c}{CIFAR-10} & \multicolumn{2}{c}{CIFAR-100} & \multicolumn{2}{c}{SVHN} & \multicolumn{2}{c}{Tiny-ImageNet} \\
\cmidrule(lr){2-3} \cmidrule(lr){4-5} \cmidrule(lr){6-7} \cmidrule(lr){8-9}
$B$ & $\sigma$ & Steps & $\sigma$ & Steps & $\sigma$ & Steps & $\sigma$ & Steps \\
\midrule
256 & 0.85 & 9750 & 0.85 & 9750 & 0.77 & 14300 & 0.72 & 19500 \\
512 & 1.04 & 4850 & 1.04 & 4850 & 0.92 & 7150 & 0.85 & 9750 \\
1024 & 1.32 & 2400 & 1.32 & 2400 & 1.15 & 3550 & 1.04 & 4850 \\
2048 & 1.74 & 1200 & 1.74 & 1200 & 1.48 & 1750 & 1.32 & 2400 \\
4096 & 2.35 & 600 & 2.35 & 600 & 1.99 & 850 & 1.74 & 1200 \\
8192 & 3.25 & 300 & 3.25 & 300 & 2.71 & 400 & 2.35 & 600 \\
\bottomrule
\end{tabular}
\end{table}

\section{Extended Tiny-ImageNet Results}
\label{sec:tinyimagenet_extended}

\begin{figure}[htbp]
\centering
\includegraphics[width=0.55\linewidth]{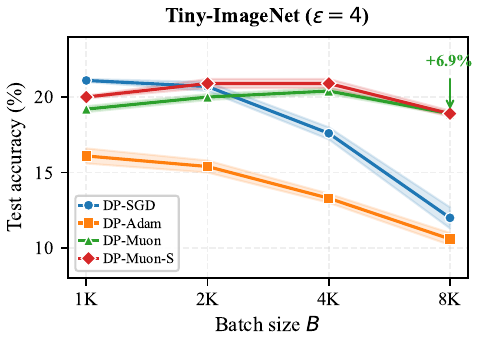}
\caption{Batch size sweep on Tiny-ImageNet ($\varepsilon = 4$, WRN-16-4). Mean $\pm$ std over three seeds. DP-SGD and DP-Adam collapse at large $B$ while spectral methods remain stable.}
\label{fig:tinyimagenet_batch}
\end{figure}

The main paper reports Tiny-ImageNet results ($64 \times 64$, 200 classes, 100K images, WRN-16-4, $\varepsilon = 4$) for batch sizes $B \in \{1024, 2048, 4096\}$.
\Cref{fig:tinyimagenet_batch,tab:tinyimagenet_full} extend this to $B = 8192$ with three-seed averages, where the contrast between spectral and non-spectral methods is most pronounced.
DP-SGD degrades from $21.1\%$ at $B = 1024$ to $12.0\%$ at $B = 8192$, approaching random performance (5\% for 200 classes).
DP-Adam follows the same collapse ($16.1\% \to 10.6\%$).
Spectral methods remain stable: DP-Muon declines only $0.3\%$ ($19.2\% \to 18.9\%$) and DP-Muon-S declines $1.1\%$ ($20.0\% \to 18.9\%$), yielding a gap of $+6.9\%$ over DP-SGD at $B = 8192$.
The low inter-seed variance of spectral methods ($\leq 0.3\%$) contrasts with DP-SGD's increasing instability at large batches ($\pm 0.7\%$ at $B = 8192$).

\begin{table}[htbp]
\centering
\caption{Test accuracy (\%) on Tiny-ImageNet at $\varepsilon = 4$ (WRN-16-4). Mean $\pm$ std over three seeds. Best per column in \textbf{bold}.}
\label{tab:tinyimagenet_full}
\begin{tabular}{l c c c c}
\toprule
Method & $B\!=\!1024$ & $B\!=\!2048$ & $B\!=\!4096$ & $B\!=\!8192$ \\
\midrule
DP-SGD    & $\mathbf{21.1 \pm 0.1}$ & $20.7 \pm 0.3$ & $17.6 \pm 0.4$ & $12.0 \pm 0.7$ \\
DP-Adam   & $16.1 \pm 0.5$ & $15.4 \pm 0.4$ & $13.3 \pm 0.3$ & $10.6 \pm 0.4$ \\
DP-Muon   & $19.2 \pm 0.2$ & $20.0 \pm 0.2$ & $20.4 \pm 0.2$ & $18.9 \pm 0.1$ \\
DP-Muon-S & $20.0 \pm 0.1$ & $\mathbf{20.9 \pm 0.3}$ & $\mathbf{20.9 \pm 0.3}$ & $\mathbf{18.9 \pm 0.2}$ \\
\bottomrule
\end{tabular}
\end{table}

\section{ImageNet-1k From-Scratch Classification}
\label{sec:imagenet_fromscratch}

Beyond the gradient-spectrum analysis of \cref{sec:imagenet_spectrum}, the spectral mechanism is evaluated on full from-scratch training of NF-ResNet-50 on ImageNet-1k ($224 \times 224$, 1000 classes), the standard large-scale setting.
All three optimizers use an identical protocol (batch size $32{,}768$, 120 epochs, augmentation multiplicity $K = 4$ applied identically), so the comparison isolates the optimizer.
\Cref{tab:imagenet_fromscratch} reports top-1 accuracy at $\varepsilon \in \{4, 8\}$ averaged over three seeds.

DP-Muon improves over DP-SGD by $+2.50\%$ at $\varepsilon = 4$ and $+2.72\%$ at $\varepsilon = 8$.
Across-seed variance is $7\times$ and $3\times$ lower respectively, indicating more reliable training and not only higher accuracy.
DP-AdamW trails both, consistent with the weak batch-size scaling of adaptive DP optimizers.
This is the standalone spectral mechanism at full scale, with no temporal-denoising components.

\begin{table}[htbp]
\centering
\caption{Top-1 accuracy (\%) for from-scratch NF-ResNet-50 on ImageNet-1k. $B = 32{,}768$, 120 epochs, $K = 4$, three seeds. Best per column in \textbf{bold}.}
\label{tab:imagenet_fromscratch}
\begin{tabular}{l c c}
\toprule
Method & $\varepsilon = 4$ & $\varepsilon = 8$ \\
\midrule
DP-SGD   & $18.47 \pm 0.41$ & $24.60 \pm 1.18$ \\
DP-AdamW & $10.37 \pm 0.21$ & $18.64 \pm 0.10$ \\
DP-Muon  & $\mathbf{20.97 \pm 0.13}$ & $\mathbf{27.32 \pm 0.16}$ \\
\bottomrule
\end{tabular}
\end{table}

\section{Compute and Memory Overhead}
\label{sec:compute_memory}

The wall-clock and memory cost of the Newton-Schulz orthogonalization is measured on WRN-16-4 ($2.7$M parameters) in fp32 on a single NVIDIA A100 (40\,GB, PCIe).
\Cref{tab:compute_overhead} reports milliseconds per logical optimization step (mean $\pm$ std over three runs) for DP-SGD and DP-Muon.

The five Newton-Schulz iterations add a roughly constant ${\sim}6.5$\,ms per step, independent of batch size and augmentation multiplicity, because orthogonalization acts on each parameter matrix rather than on each sample.
Per-sample backpropagation scales with $B \cdot K$, so the relative overhead shrinks as the workload grows.
It falls from $+2.92\%$ at $B = 2048, K = 1$ to $+0.77\%$ at $B = 4096, K = 4$.
Peak GPU memory is within $1\%$ across optimizers at each configuration.
DP-Muon maintains a single first-order momentum buffer ($1.0\times$ model size), versus DP-Adam's two ($2.0\times$, first and second moments).

\begin{table}[htbp]
\centering
\caption{Wall-clock overhead of Newton-Schulz orthogonalization. WRN-16-4, fp32, single A100. Milliseconds per logical step, mean $\pm$ std over three runs. $\Delta$ is the relative overhead of DP-Muon over DP-SGD.}
\label{tab:compute_overhead}
\begin{tabular}{l c c c}
\toprule
Configuration & DP-SGD & DP-Muon & $\Delta$ \\
\midrule
$B = 2048,\ K = 1$ & $488.1 \pm 1.1$ & $502.4 \pm 0.8$ & $+2.92\%$ \\
$B = 4096,\ K = 4$ & $3894.6 \pm 1.0$ & $3924.4 \pm 11.1$ & $+0.77\%$ \\
\bottomrule
\end{tabular}
\end{table}

\end{document}